\documentclass[10pt,twocolumn,letterpaper]{article}

\usepackage{cvpr}
\usepackage{times}
\usepackage{epsfig}
\usepackage{graphicx}
\usepackage{amsmath}
\usepackage{amssymb}

\usepackage{booktabs}
\usepackage{xcolor,colortbl}
\usepackage[caption=false]{subfig}

\usepackage[breaklinks=true,bookmarks=false]{hyperref}

\cvprfinalcopy 

\def\cvprPaperID{****} 

\ifcvprfinal\fi
\begin{document}

\title{Accel: A Corrective Fusion Network for \\ Efficient Semantic Segmentation on Video}

\author{Samvit Jain, Xin Wang, Joseph Gonzalez\\
\\[-8pt]
University of California, Berkeley\\
{\tt\small \{samvit,xinw,jegonzal\}@eecs.berkeley.edu}
}

\maketitle

\begin{abstract}
	We present Accel, a novel semantic video segmentation system that achieves high accuracy at low inference cost by combining the predictions of two network branches: (1) a \textit{reference branch} that extracts high-detail features on a reference keyframe, and warps these features forward using frame-to-frame optical flow estimates, and (2) an \textit{update branch} that computes features of adjustable quality on the current frame, performing a temporal update at each video frame. The modularity of the update branch, where feature subnetworks of varying layer depth can be inserted (e.g. ResNet-18 to ResNet-101), enables operation over a new, state-of-the-art accuracy-throughput trade-off spectrum. Over this curve, Accel models achieve both higher accuracy and faster inference times than the closest comparable single-frame segmentation networks. In general, Accel significantly outperforms previous work on efficient semantic video segmentation, correcting warping-related error that compounds on datasets with complex dynamics. Accel is end-to-end trainable and highly modular: the reference network, the optical flow network, and the update network can each be selected independently, depending on application requirements, and then jointly fine-tuned. The result is a robust, general system for fast, high-accuracy semantic segmentation on video.
\end{abstract}

\section{Introduction}

Semantic segmentation is an intensive computer vision task that involves generating class predictions for each pixel in an image, where classes range from foreground objects such as ``person'' and ``vehicle'' to background entities such as ``building''  and ``sky''. When applied to frames in high resolution video, this task becomes yet more expensive, as the high spatial dimensionality of the output is further scaled by the video's temporal frame rate (e.g. 30 frames per second). By treating video as a collection of uncorrelated still images, contemporary approaches to semantic video segmentation incur this full computational cost, achieving inference throughput of less than 1.5 frames per second (fps) on a 30 fps video feed \cite{DeepLab,DCN,DRN}. Moreover, by ignoring temporal context, frame-by-frame approaches fail to realize the potential for improved accuracy offered by the availability of preceding frames in a scene.

\begin{figure}[t]
	\centering
	\vspace{-8mm}
	\includegraphics[width=14cm]{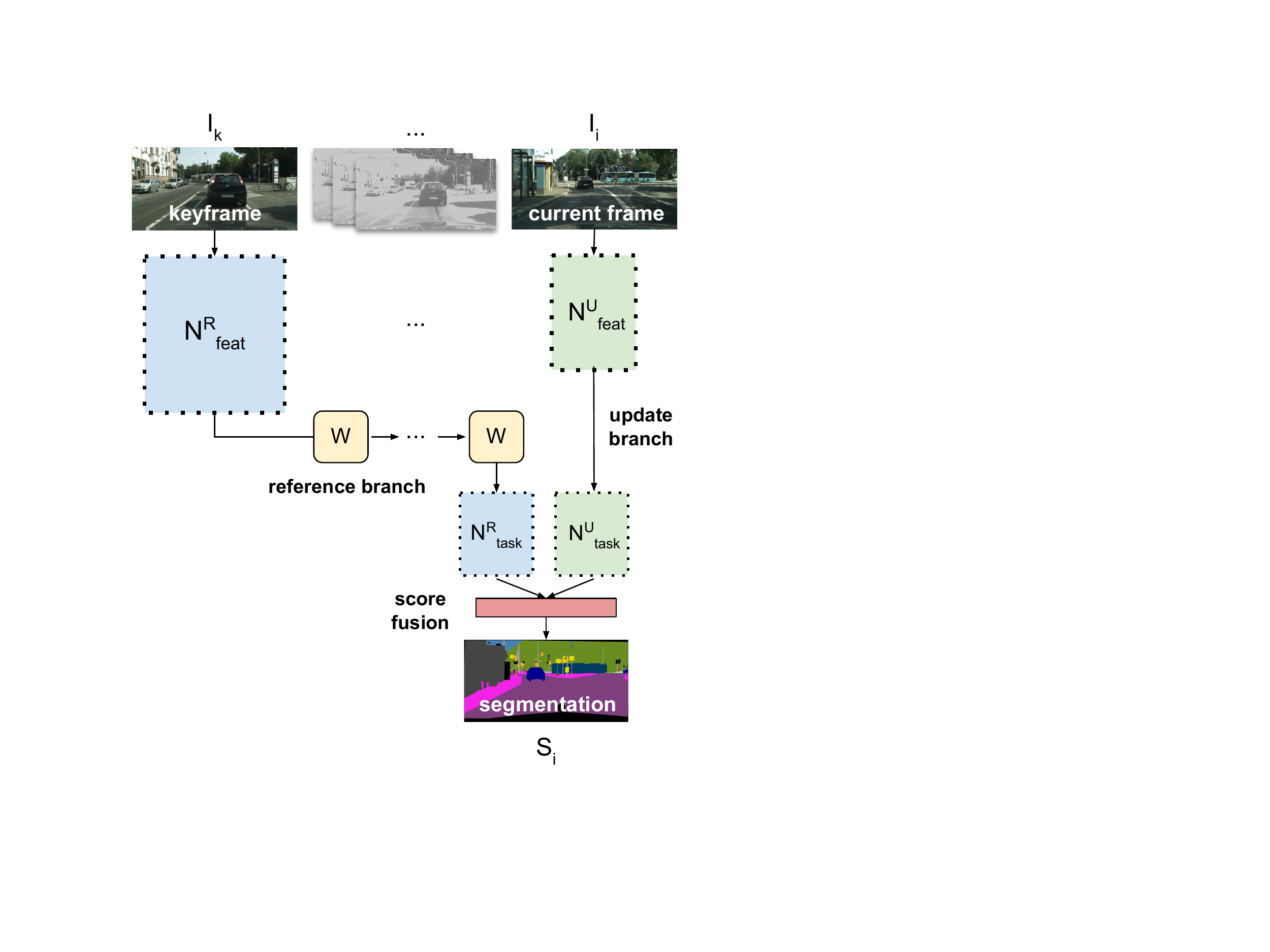}
	\vspace{-25mm}
	\caption{Accel is a fast, high-accuracy, end-to-end trainable video recognition system that combines two network branches: 1) a \textit{reference branch} that computes a score map on high-detail features warped from the last visited keyframe, and 2) a cheap \textit{update branch} that corrects this prediction based on features of adjustable quality (e.g. ResNet-18 to -101) computed on the current frame.}
	\label{fig:abstract}
	\vspace{-4mm}
\end{figure}

Prior work has proposed feature reuse and feature warping as means to reduce computation on video. In particular, exploiting the observation that higher-level representations evolve more slowly than raw pixels in a video \cite{CC}, these approaches relegate feature extraction, the most expensive component of most video recognition architectures \cite{DFF}, to select keyframes, and project these features forward via na\"ive copying or warping based on optical flow. While feature warping does enable some speedup \cite{DFF}, its efficacy is constrained by video dynamics. Fast scene evolution necessitates frequent feature re-computation, and feature warping in videos with a moving observer (e.g. driving footage), where the entire scene moves relative to the camera, introduces significant warping error. Warping error compounds with repeated application of the warping operator.

Our proposed system, Accel (Fig. \ref{fig:abstract}), addresses the challenges of efficient video segmentation by combining the predictions of a \textit{reference branch}, which maintains an incrementally warped representation of the last keyframe, with the predictions of an \textit{update branch}, which processes the current frame, in a convolutional fusion step. Importantly, this \textit{update branch} has the ability to serve two purposes: 1) correction and 2) anchoring. When a cheap, shallow update network is used (e.g. ResNet-18), the warped keyframe features form the more accurate input to the fusion operator, and the update branch \textit{corrects} warping-related error with information from the current frame. When an expensive, deep update network is used (e.g. ResNet-101), the update branch \textit{anchors} the network on the features of the current frame, which is the higher accuracy input, while the reference branch augments the prediction with context from preceding frames. These two modes of operation represent two extremes on the highly competitive accuracy-throughput trade-off curve Accel unlocks.

We evaluate Accel on Cityscapes and CamVid, the largest available video segmentation datasets \cite{CamVid,KITTI,Cityscapes}, and demonstrate a full range of accuracy-inference speed modalities. Our reference network, which we operate on keyframes, is an implementation of the DeepLab segmentation architecture \cite{DeepLab} based on ResNet-101. Our chosen update networks range from the fast ResNet-18 (in Accel-18) to the accurate ResNet-101 (in Accel-101). On the high throughput side, the cheapest version of Accel, Accel-18, is both faster and more accurate than the closest comparable DeepLab model. On the high accuracy side, Accel-101 is more accurate than the best available single-frame model, DeepLab-101. As a set, the ensemble of Accel models achieve significantly higher accuracy than previous work on the problem at every keyframe interval. Taken togther, these results form a new state-of-the-art on the task of efficient semantic video segmentation.

\section{Related Work}

\subsubsection{Image Semantic Segmentation}

Semantic video segmentation is a recent offshoot of the study of semantic \textit{image} segmentation, a problem of long-standing interest in computer vision. The classical approach to image segmentation was to propagate information about pixel assignments through a graphical model \cite{EGB,Texton,EHG}, a costly technique that scaled poorly to complex image datasets \cite{GEP}. Most recent research follows the lead of Long et al. in the use of fully convolutional networks (FCNs) to segment images \cite{FCN}. Recent work has augmented the FCN model with explicit encoder-decoder architectures \cite{SegNet,RefineNet}, dilated convolutions \cite{Multiscale,DRN}, and post-processing CRFs \cite{DeepLabV1,DeepLab}, achieving higher accuracy on larger, more realistic datasets \cite{CamVid,PascalVOC,Cityscapes}.

\subsubsection{Video Semantic Segmentation}

Unlike video object segmentation, where a vast literature exists on using motion and temporal cues to track and segment objects across frames \cite{FastObjSeg,MovingObjects,FewStrokes,ObjectFlow}, the video \textit{semantic} segmentation task, which calls for a pixel-level labeling of the entire frame, is less studied. The rise of applications in autonomous control and video analysis, along with increased concern about the acute computational cost of na\"ive frame-by-frame approaches, however, have sparked significant interest in the problem of \textit{efficient video segmentation}. Recent papers have proposed selective re-execution of feature extraction layers \cite{CC}, optical flow-based feature warping \cite{DFF}, and LSTM-based, fixed-budget keyframe selection policies \cite{Budget} as means to achieve speedup on video. Of the three, the optical flow-based approach \cite{DFF} is the strongest contender, achieving greater cost savings and higher accuracy than both the first approach, which na\"ively copies features, and the third, which is \textit{offline} and has yet to demonstrate strong quantitative results. Despite its relative strength, however, flow-based warping \cite{DFF} introduces compounding error in the intermediate representation, and fails to incorporate other forms of temporal change (e.g. new objects, occlusions). As a result, significant accuracy degradation is observed at moderate to high keyframe intervals, restricting its achievable speedup.

To address these problems, new work has proposed adaptive feature propagation, partial feature updating, and adaptive keyframe selection as schemes to optimally schedule and propagate computation on video \cite{HighPerfVOD,LowLatency,DynamicVSN}. These techniques have the drawback of complexity, requiring the network to learn auxiliary representations to decide: (1) whether to recompute features for a region or frame, and (2) how to propagate features in a spatially-variant manner. Moreover, they do not fundamentally address the problem of mounting warping error, instead optimizing the \textit{operation} of \cite{DFF}. In contrast, in Accel, we resolve the challenges by proposing a simple network augmentation: a second branch that cheaply processes each video frame, and corrects accumulated temporal error in the reference representation.

\subsubsection{Network Fusion}

Feature and network fusion have been extensively explored in other contexts. A body of work, beginning with \cite{TwoStream} and extending to \cite{TwoStream2,STResNets,DeepReps}, studies spatial and temporal two-stream fusion for video action recognition. In the two-stream model, softmax scores of two network branches, one which operates on single RGB frames (spatial stream) and another on multi-frame optical flow fields (temporal stream), are fused to discern actions from still video frames. Variants of this approach have been subsequently applied to video classification \cite{Karpathy,Zuxuan} and video object segmentation \cite{FusionSeg,Tokmakov}, among other tasks. Unlike spatio-temporal fusion, which attempts to jointly deduce scene structure from RGB frames and motion for video-level tasks, the Accel fusion network uses keyframe context and optical flow as a means to conserve computation and boost accuracy in intensive \textit{frame} and \textit{pixel-level prediction tasks}, such as segmentation. In Accel, both branches process representations of single frames, and motion (optical flow) is used implicitly in the model to update a latent reference representation. Together, these design choices make Accel robust and configurable. The fact that network components are independent, with clear interfaces, allows the entire system to be operated at multiple performance modalities, via choice of update network (e.g. ResNet-$x$), motion input (e.g. optical flow, H.264 block motion \cite{BMV}), and keyframe interval.

\section{Approach}

\subsection{Problem Statement}

Given a video $I$ composed of frames $\{I_1, I_2, ... I_T\}$, we wish to compute the segmentation of each frame: $\{S_1, S_2, ... S_T\}$. We have at our disposal a single-frame segmentation network $N$ that can segment any still frame in the video: $N(I_i) = S_i$. This network is accurate, but slow. Since $N$ only takes single images as input, it cannot exploit the temporal continuity of video; the best we can do is to run $N$ on every frame $I_i \in I$ sequentially.

Instead, we would like to develop a \textit{video} segmentation network $N'$ that takes as input a frame $I_i$, and potentially additional context (e.g. nearby frames, features, or segmentations), and renders $S'_i$. Our goals are two-fold: (1) $\{S'_i\}$ should be \textit{at least as accurate} as $\{S_i\}$, and (2) running $N'(\{I_i\})$ should be \textit{faster} than running $N(\{I_i\})$.

\subsection{Operation Model}

Our base single-frame semantic segmentation architecture $N$ consists of three functional components: (1) a feature subnetwork $N_{feat}$ that takes as input an RGB image $I_i \in R^{1 \times 3 \times h \times w}$ and returns an intermediate representation $f_i \in R^{1 \times 2048 \times \frac{h}{16} \times \frac{w}{16}}$, (2) a task subnetwork $N_{task}$ that takes as input the intermediate representation $f_i$ and returns a semantic segmentation score map $s_i \in R^{1 \times C \times h \times w}$, where $C$ is the number of labeled classes in the dataset, and (3) an output block $P$ that converts $s_i$ to normalized probabilities $p_i \in [0, 1]^{1 \times C \times h \times w}$ and then segmentations $S_i \in R^{1 \times 1 \times h \times w}$.

This division follows a common pattern in image and video recognition architectures \cite{DFF}. The feature network, $N_{feat}$, is largely identical across different recognition tasks (object detection, instance segmentation, semantic segmentation), and is obtained by discarding the final $k$-way classification layer in a standard image classification network (e.g. ResNet-101), and decreasing the stride length in the first block of the conv5 layer from 2 to 1 to obtain higher-resolution feature maps (spatial dimension $\frac{h}{16} \times \frac{w}{16}$ instead of $\frac{h}{32} \times \frac{w}{32}$). The task network, $N_{task}$, for semantic segmentation includes three blocks: (1) a feature projection block, which consists of a $1 \times 1$ convolution, plus a non-linear activation (ReLU), and reduces the feature channel dimension from 2048 to 1024, (2) a scoring layer, which consists of a single $1 \times 1$ convolution, and further reduces the channel dimension from 1024 to the $C$ semantic classes, and (3) an upsampling block, which consists of a deconvolutional layer and a cropping layer, and upsamples the predicted scores from $\frac{h}{16} \times \frac{w}{16}$ to the spatial dimensionality of the input image, $h \times w$. Finally, output block $P$ consists of a softmax layer, followed by an argmax layer.

Exploiting the observation that features can be reused across frames to reduce computation \cite{CC,DFF}, we now adopt the following operation model on video. $N_{feat}$, which is deep and expensive, is executed only on select, designated \textit{keyframes}. Keyframes are selected at regular intervals, starting with the first frame in the video. The extracted keyframe features $f_i$ are warped to subsequent frames using a computed optical flow field, $O$. $N_{task}$, which is shallow and cheap, is executed on every frame. Since computing optical flow $O(I_i, I_j)$ on pairs of frames, and warping features with the flow field $W(f_i, O)) \rightarrow \hat{f_j}$, is much cheaper than computing $N_{feat}(I_j)$ \cite{DFF}, this scheme saves significant computation on \textit{intermediate frames}, which form the vast majority of video frames.

\subsection{Accel}

\begin{figure}[t]
	\centering
	\vspace{-10mm}
	\includegraphics[width=11.2cm]{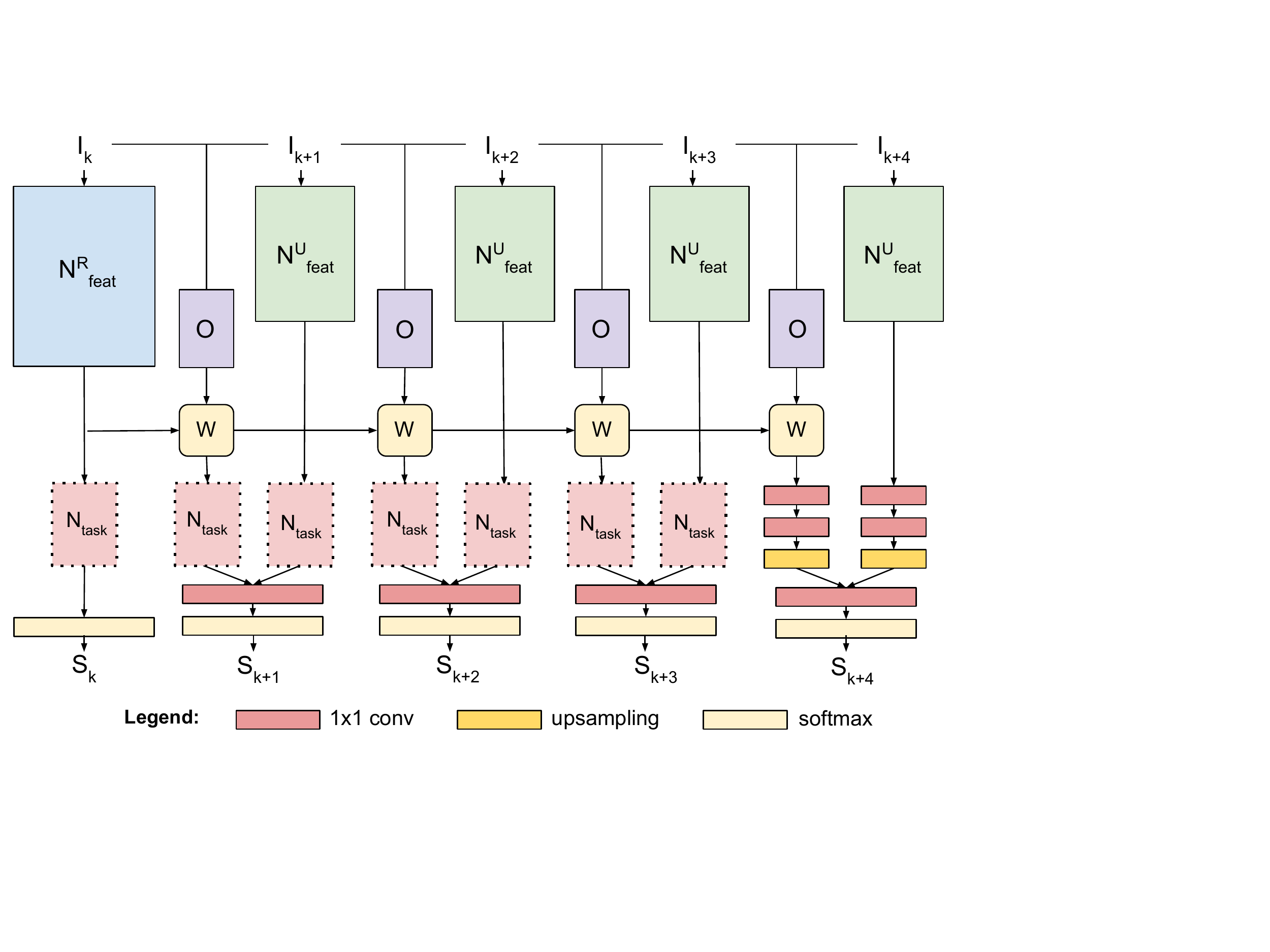}
	\vspace{-22mm}
	\caption{Accel consists of several components: (1) a reference feature net $N^R_{feat}$ executed on keyframes, (2) an update feature net $N^U_{feat}$ executed on intermediate frames, (3) an optical flow net $O$ used for feature warping $W$, (4) two instantiations of $N_{task}$ (reference and update), (5) a $1 \times 1$ conv network fusion layer, and (6) a final softmax layer.}
	\label{fig:accel}
	\vspace{-2mm}
\end{figure}

In Accel, we introduce a lightweight feature network, $N_{feat}^U$, on intermediate frames to update score predictions based on the warped keyframe features, with information from the current frame. On keyframes, we execute our original feature network, now denoted as the \textit{reference} feature network, $N_{feat}^R$. In our system, we use ResNet-101 as $N_{feat}^R$, and a range of models, from ResNet-18 to ResNet-101, as $N_{feat}^U$, depending on specific accuracy-performance goals. In this section, we discuss a forward pass through this new architecture, Accel (see Fig. \ref{fig:accel}).

On keyframes, denoted by index $k$, we execute the full reference network $P(N_{task}^R(N_{feat}^R(I_k)))$ to yield a segmentation $S_k$, and save the intermediate output $f_k = N_{feat}^R(I_k)$ as our cached features $f^c$.

On intermediate frames $i$, we compute scores $s_i^R$ and $s_i^U$ along both a \textit{reference branch} and an \textit{update branch}, respectively. On the reference branch, we warp $f^c$ from the previous frame $I_{i-1}$ to the current frame $I_i$, and then execute $N_{task}^R$. As our warping operation $W$, we spatially transform our cached features $f^c$ with a bilinear interpolation of the optical flow field $O(I_{i-1}, I_i)$, as in \cite{DFF}. On the update branch, we run the full update network $N^U$. Symbolically, the two branches can be represented as:
\begin{align}
s_i^R &= N_{task}^R(W(f^c, O(I_{i-1}, I_i))) \\
s_i^U &= N_{task}^U(N_{feat}^U(I_i))
\end{align}

The score maps $s_i^R$ and $s_i^U$ represent two views on the correct class labels for the pixels in the current frame. These predictions are now merged in a $1 \times 1$ convolutional fusion step, which we refer to as \textit{score fusion} (SF). $s_i^R$ and $s_i^U$ are stacked along the channel dimension, yielding an input $s_i^{stacked} \in R^{1 \times 2C \times h \times w}$. Applying a $1 \times 1$ convolutional layer with dimensions $C \times 2C \times 1 \times 1$ to $s_i^{stacked}$ yields an output $s_i \in R^{1 \times C \times h \times w}$. Notationally, $s_i = SF(s_i^{stacked}) = SF([s_i^R, s_i^U])$. Finally, applying the output block $P$ to $s_i$ yields the segmentation $S_i$ of frame $I_i$.

Note that while the layer definitions of $N_{feat}^R$ and $N_{feat}^U$ differ in general, $N_{task}^R$ and $N_{task}^U$ are architecturally equivalent, albeit independent instantiations (i.e. they don't share weights). This makes Accel highly modular. Since the task network $N_{task}$ has a fixed interface, Accel can accept any feature network $N_{feat}^U$ that outputs representations $f_i$ with the appropriate dimensionality.

\subsection{Training} \label{sec:approach-train}

Accel can be trained end-to-end on sparsely annotated sequences of video frames. The entire network consists of the score fusion layer, along with three independently trainable components, $N^R$, $N^U$, and $O$, which we now discuss.

For our reference network $N^R$ and update network $N^U$, we use a high-accuracy variant \cite{DCN} of the DeepLab architecture \cite{DeepLab}. DeepLab is a canonical architecture for semantic segmentation \cite{DCN,SegNet,RefineNet,Multiscale}, and a DeepLab implementation has consistently ranked first in the Pascal VOC segmentation challenge \cite{Pascal}. $N_{feat}^R$ and $N_{feat}^U$ are first trained on ImageNet; $N^R$ and $N^U$ are then individually fine-tuned on a semantic segmentation dataset, such as Cityscapes \cite{Cityscapes}. In Accel, we fix $N_{feat}^R$ as ResNet-101. We then build an ensemble of models, based on a range of update feature networks $N_{feat}^U$: ResNet-18, -34, -50, and -101. This forms a full spectrum of accuracy-throughput modalities, from a lightweight, competitive Accel based on ResNet-18, to a slow, extremely accurate Accel based on ResNet-101. For the third and last independently trainable component, the optical flow network $O$, we use the ``Simple" architecture from the FlowNet project \cite{FlowNet}. This network is pre-trained on the synthetic Flying Chairs dataset, and then jointly fine-tuned on the semantic segmentation task with $N^R$.

To train Accel, we initialize with weights from these three pre-trained models. In each mini-batch, we select a frame $I_{j}$. When training at keyframe interval $n$, we select frame $I_{j-(n-1)}$ from the associated video snippet, and mark it as the corresponding keyframe $I_k$ for frame $I_{j}$. In a forward pass, we execute Accel's reference branch on frame $I_k$, and execute the update branch and fusion step on each subsequent intermediate frame until $I_j$. A pixel-level, cross-entropy loss \cite{FCN} is computed on the predicted segmentation $S_j$ and the ground-truth label for frame $I_{j}$. In the backward pass, gradients are backpropagated through time through the score fusion operator, the reference and update branches, and the warping operator, which is parameter-free but fully differentiable. Note that the purpose of joint training is to learn weights for the score fusion (SF) operator, and to optimize other weights (i.e. $N_{task}^R$ and $N_{task}^U$) for the end-to-end task.

\subsection{Design Choices}

Recent work has explored \textit{adaptive keyframe scheduling}, where keyframes are selected based on varying video dynamics and feature quality \cite{HighPerfVOD,LowLatency,DynamicVSN}. Here both rapid scene change and declining feature quality can trigger feature recomputation. In practice, we find these techniques introduce significant complexity, without yielding a commensurate gain in accuracy. Note that keyframe scheduling is an optimization that is orthogonal to network design, and therefore compatible with Accel.

We also note that other segmentation architectures exist that cannot be as easily divided into an expensive $N_{feat}$ and a cheap $N_{task}$. These include the highly accurate pyramid spatial pooling networks (PSPNet, NetWarp) \cite{PSPNet,NetWarp}, symmetric encoder-decoder style architectures (U-Net) \cite{UNet}, and the parameter efficient, fully-convolutional DenseNets (FC-DensetNet) \cite{FCDenseNet}. Each, however, posses a design feature which make them unsuitable for the problem of \textit{video} semantic segmentation.
PSPNet and NetWarp are exceedingly slow, at over 3.0 seconds per frame on Cityscapes (more than \textit{four times slower} than DeepLab-101) \cite{NetWarp}. U-Net was built for the precise segmentation of biomedical structures, where training data is extremely sparse, and optimizes for this domain. Finally, FC-DenseNet, which utilizes dense feed-forward connections between every pair of layers to learn representations more effectively, has yet to demonstrate strong quantitative results on large datasets.

\section{Experiments}

\subsection{Setup}

We evaluate Accel on Cityscapes \cite{Cityscapes} and CamVid \cite{CamVid}, the largest available datasets for complex urban scene understanding and the standard benchmarks for semantic video segmentation \cite{DeepLab,DCN,DRN}. Cityscapes consists of 30-frame snippets of street scenes from 50 European cities, recorded at a frame rate of 17 frames per second (fps). Individual frames are $2048 \times 1024$ pixels in size. The train, validation, and test sets consist of 2975, 500, and 1525 snippets each, with ground truth labels provided for the 20\textsuperscript{th} frame in each snippet in the train and validation set. The Cambridge-Driving Labeled Video Database (CamVid) consists of over 10 minutes of footage captured at 30 fps. Frames are 960 by 720 pixels in size, and ground-truth labels are provided for every 30\textsuperscript{th} frame. We use the standard train-test split of \cite{Sturgess}, which divides CamVid into three train and two test sequences, containing 367 and 233 frames, respectively.

To evaluate accuracy, we use the \textit{mean intersection-over union }(mIoU) metric, standard for semantic segmentation \cite{PascalVOC}. mIoU is defined as the average achieved intersection-over-union value, or Jaccard index, over all valid semantic classes in the dataset. To evaluate performance, we report \textit{average inference time} in seconds per frame (s/frame) over the entire dataset. Note that this is the inverse of throughput (frames per second).

We train Accel as described in Section: Approach on Cityscapes and CamVid. We perform 50 epochs of joint training at a learning rate of $5 \cdot 10^{-4}$ in two phases. In phase one, all weights except SF are frozen. In phase two, after 40 epochs, all remaining weights are unfrozen. We train a reference implementation of \cite{DFF} by jointly fine-tuning the same implementations of $N^R$ and $O$. At inference time, we select an operational keyframe interval $i$, and in each snippet, choose keyframes such that the distance to the labeled frame rotates uniformly through $[0, i-1]$. This sampling procedure simulates evaluation on a densely labeled video dataset, where $\frac{1}{i}$ frames fall at each keyframe offset between $0$ and $i-1$. Here we follow the example of previous work \cite{DFF}.

Finally, Accel is implemented in the MXNet framework \cite{MXNet}. All experiments are run on Tesla K80 GPUs, at keyframe interval 5, unless otherwise stated. Our implementation of Accel is open-source on GitHub.

\subsection{Results}

\subsubsection{Baselines}

To generate our baseline accuracy-throughput curve, we run single-frame DeepLab \cite{DeepLab} models based on ResNet-18, -34, -50, and -101 on the Cityscapes and CamVid test data. For both DeepLab and Accel, we use a variant of the ResNet architecture called Deformable ResNet, which employs \textit{deformable convolutions} in the last ResNet block (conv5) to achieve significantly higher accuracy at slightly higher inference cost \cite{DCN}. We refer to DeepLab models based on ResNet-$x$ as DeepLab-$x$, and Accel models based on a ResNet-$x$ update network as Accel-$x$.

\subsubsection{Accuracy-throughput}

\begin{figure}[t]
	\centering
	\includegraphics[width=8cm]{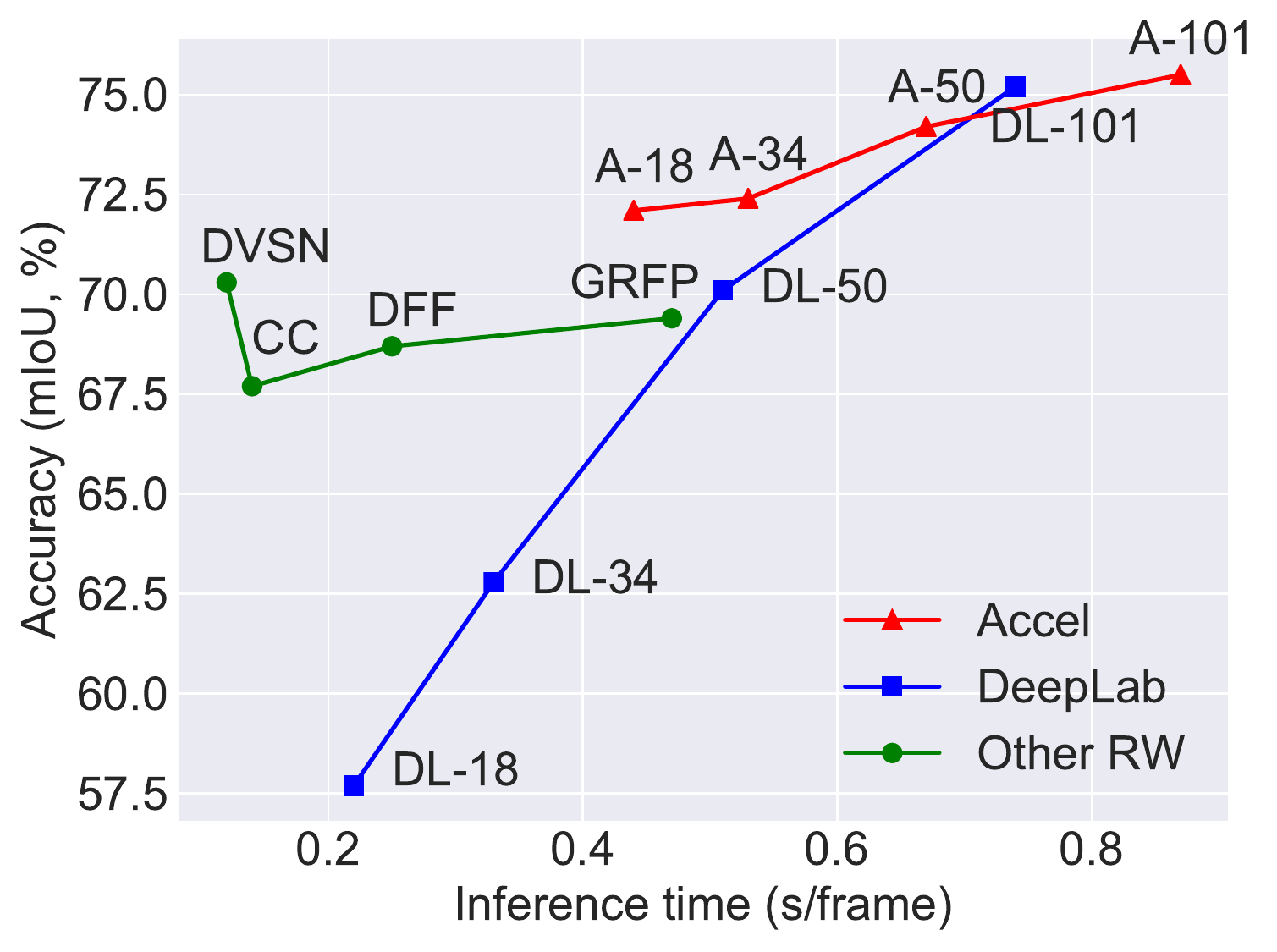}
	\vspace{-1mm}
	\caption{Accuracy vs. inference time on \textbf{Cityscapes}. Comparing four variants of Accel (A-$x$) to single-frame DeepLab models (DL-$x$) and various other related work (RW). All results at keyframe interval 5. Data from Table \ref{tbl:baselines-city}.}
	\label{fig:baselines-city}
	\vspace{-2mm}
\end{figure}

\begin{figure}[t]
	\centering
	\includegraphics[width=8cm]{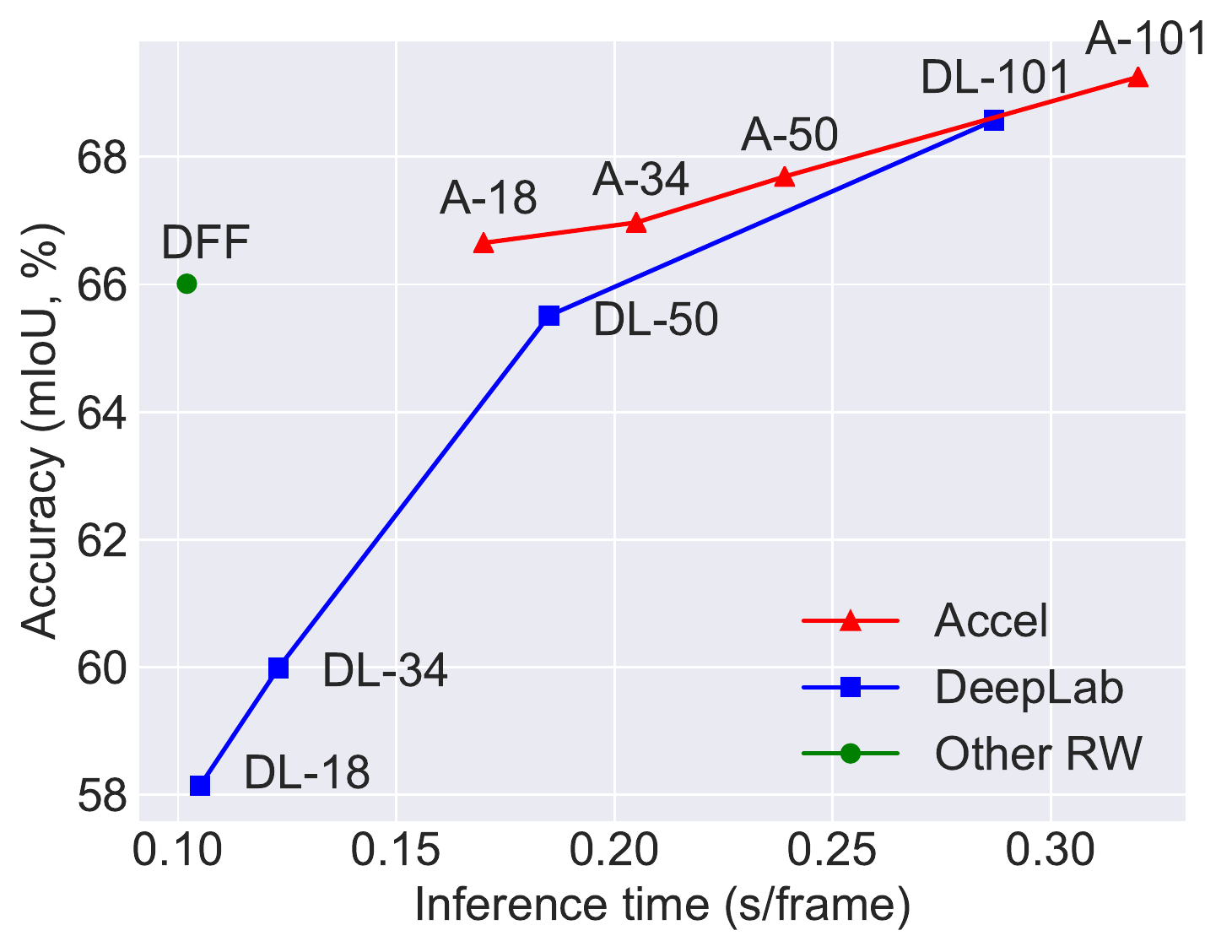}
	\vspace{-1mm}
	\caption{Accuracy vs. inference time on \textbf{CamVid}. All results at keyframe interval 5. Data from Table \ref{tbl:baselines-cam}. (CC and DVSN do not evaluate on CamVid; GRFP does not report timing results.)}
	\label{fig:baselines-cam}
	\vspace{-2mm}
\end{figure}

Using Accel, we achieve a new, state-of-the-art accuracy-throughput trade-off curve for semantic video segmentation (see Figs. \ref{fig:baselines-city}, \ref{fig:baselines-cam}).

All Accel models, from Accel-18 to Accel-101, allow operation at high accuracy: above 72 mIoU on Cityscapes and above 66 mIoU on CamVid. At the high accuracy end, Accel-101 is by far the most accurate model, achieving higher mIoU than the best available DeepLab model, DeepLab-101. At the high throughput end, Accel-18 is \textit{both faster and more accurate} than the closest comparable single-frame model, DeepLab-50. Notably, Accel-18 is over 40\% cheaper than DeepLab-101, at only 2-3\% lower mIoU. As a rule, each Accel-$x$ model is more accurate than its single-frame counterpart, DeepLab-$x$, for all $x$.

Together, the four Accel models form an operational Pareto curve that clearly supersedes the Pareto curve defined by the four single-frame DeepLab models (Figs. \ref{fig:baselines-city}, \ref{fig:baselines-cam}). Accel also visibly outperforms related work, including Clockwork Convnets \cite{CC}, Deep Feature Flow \cite{DFF}, Gated Recurrent Flow Propagation \cite{GRFP}, and Dynamic Video Segmentation Network \cite{DynamicVSN} (see Table \ref{tbl:baselines-city}). Though DFF offers a strong accuracy-throughput trade-off, due to its fixed architecture, it is not a contender in the high accuracy regime. In the next section, we compare more closely to DFF.

\renewcommand{\arraystretch}{1.2}
\begin{table}[ht]
	\vspace{+2mm}
	\caption{Accuracy and inference times on \textbf{Cityscapes} for four single-frame DeepLab models (DL-$x$), four variants of Accel (A-$x$), and various related work. Table ordered by accuracy. All inference time standard deviations less than 0.01. Each Accel-$x$ model is more accurate than its single-frame counterpart, DeepLab-$x$, for all $x$. Data plotted in Fig. \ref{fig:baselines-city}.}
	\centering
	\label{tbl:baselines-city}
	\begin{tabular}[t]{@{\extracolsep{4pt}}lcc}
		\toprule
		Model & Acc. (mIoU, \%) & Time (s/frame) \\
		\midrule
		DL-18 & 57.7 & 0.22 \\
		DL-34 & 62.8 & 0.33 \\
		CC (Shel. 2016) & 67.7 & 0.14 \\
		DFF (Zhu 2017) & 68.7 & 0.25 \\
		GRFP (Nils. 2018) & 69.4 & 0.47 \\
		DL-50 & 70.1 & 0.51 \\
		DVSN (Xu 2018) & 70.3 & 0.12 \\
		\textbf{A-18} & 72.1 & 0.44 \\
		\textbf{A-34} & 72.4 & 0.53 \\
		\textbf{A-50} & 74.2 & 0.67 \\
		DL-101 & 75.2 & 0.74 \\
		\textbf{A-101} & 75.5 & 0.87 \\
		\bottomrule
	\end{tabular}
\end{table}
\renewcommand{\arraystretch}{1.0}

\renewcommand{\arraystretch}{1.2}
\begin{table}[ht]
	\vspace{+2mm}
	\caption{Accuracy and inference times on \textbf{CamVid}. Table ordered by accuracy. Data plotted in Fig. \ref{fig:baselines-cam}.}
	\centering
	\label{tbl:baselines-cam}%
	\begin{tabular}[t]{@{\extracolsep{4pt}}lcc}
		\toprule
		Model & Acc. (mIoU, \%) & Time (s/frame) \\
		\midrule
		DL-18 & 58.1 & 0.105 \\
		DL-34 & 60.0 & 0.123 \\
		DL-50 & 65.5 & 0.185 \\
		DFF (Zhu 2017) & 66.0 & 0.102 \\
		\textbf{A-18} & 66.7 & 0.170 \\
		\textbf{A-34} & 67.0 & 0.205 \\
		\textbf{A-50} & 67.7 & 0.239 \\
		DL-101 & 68.6 & 0.287 \\
		\textbf{A-101} & 69.3 & 0.320 \\
		\bottomrule
	\end{tabular}
	\vspace{-2mm}
\end{table}
\renewcommand{\arraystretch}{1.0}

\subsubsection{Keyframe intervals}

\begin{figure}[t]
	\centering
	\vspace{-0mm}
	\includegraphics[width=8cm]{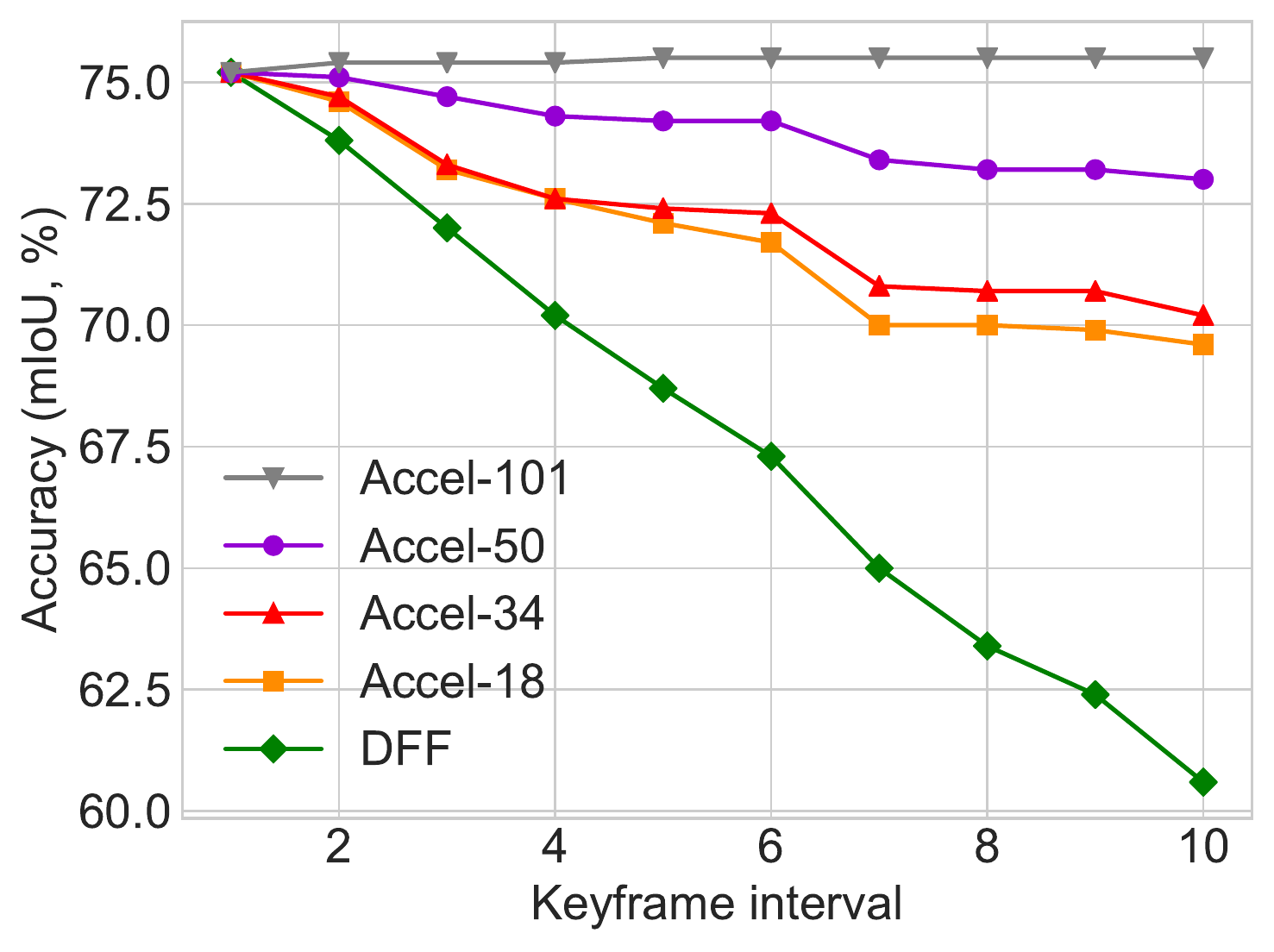}
	\vspace{-1mm}
	\caption{Accuracy vs. keyframe interval on \textbf{Cityscapes} for optical flow-based warping alone (DFF) and four variants of Accel. All five schemes use ResNet-101 in $N^R$.}
	\label{fig:keyframe}
	\vspace{-2mm}
\end{figure}

In this section, we extend our evaluation to a range of keyframe intervals from $1$ to $10$. Keyframe interval $1$ corresponds to running the reference network $N^R$ on every frame. As a result, Deep Feature Flow (DFF) \cite{DFF} and the Accel variants report the same accuracy at this setting (see Fig. \ref{fig:keyframe}). At keyframe intervals above $1$, we find that even the cheapest version of Accel, Accel-18, consistently offers higher accuracy than DFF. In particular, over keyframe interval 8, a wide accuracy gap emerges, as DFF's accuracy approaches $60$ mIoU while all Accel models maintain roughly between $70$ and $75$ mIoU (Fig. \ref{fig:keyframe}).

This gap is an illustration of the compounding warping error that builds in DFF, but is corrected in Accel with the advent of the update branch. The trade-off is that Accel models are slower on intermediate frames: in addition to the inference cost of $O$ and $N^R_{task}$, which is also paid by DFF, Accel models also incur the cost of $N^U$, which is low when $N^U_{feat}$ is ResNet-18 and higher when $N^U_{feat}$ is ResNet-101.

\subsubsection{Ablation study}

We now present a simple ablation study that isolates the contributions of the reference network $N^R$ and the update network $N^U$ to the accuracy of Accel (see Table \ref{tbl:ablation}). Disabling $N^U$ corresponds to using only the optical flow-warped representations from the previous keyframe. Since all versions of Accel share the same $N^R$, this results in the same accuracy for all models (row 1). Disabling the reference network $N^R$ corresponds to running only the single-frame update networks, DeepLab-18, -34, -50, or -101, on all frames (row 2). Disabling neither yields our original models (row 3). Notice the effect of the network fusion: each unmodified Accel model is more accurate than either of its component subnetworks. Moreover, Accel-18 observes a 6.8 point accuracy boost over $N^R$ via the use of an update network $N^U$ that is cheaper and \textit{much less accurate} than $N^R$. This confirms the powerful synergistic effect of combining two contrasting sets of representations: one that is high-detail but dated, and one that is lower resolution but temporally current.

\renewcommand{\arraystretch}{1.1}
\begin{table}[ht]
	\vspace{+2mm}
	\caption{\textbf{Ablation study}. A breakdown of the accuracy contributions of $N^R$ (reference branch) and $N^U$ (update branch) to Accel. Results for keyframe interval $i=5$, at the max offset ($4$) from the keyframe. Cityscapes dataset.}
	\vspace{3mm}
	\centering
	\label{tbl:ablation}
	\setlength{\tabcolsep}{4.9pt}
	\begin{tabular}{@{\extracolsep{4pt}}lcccc}
		\toprule
		{} & \multicolumn{4}{c}{Model} \\
		\cmidrule{2-5}
		Setting & A-18 & A-34 & A-50 & A-101 \\
		\midrule
		$N^R$ only & 62.4 & 62.4 & 62.4 & 62.4 \\
		$N^U$ only & 57.7 & 62.8 & 70.1 & 75.2 \\
		Accel & \textbf{69.2} & \textbf{69.7} & \textbf{73.0} & \textbf{75.5} \\
		\bottomrule
	\end{tabular}
	\vspace{-2mm}
\end{table}
\renewcommand{\arraystretch}{1.0}

\subsubsection{Fusion location}

In this section, we evaluate the impact of fusion location on final network accuracy and performance. Accel, as described so far, uses a $1 \times 1$ convolutional layer to fuse pre-softmax class scores, but it was also possible to perform this fusion at an earlier stage. In Table \ref{tbl:fusion}, we compare accuracy values and inference times for two fusion variants: (1) feature fusion (fusion between $N_{feat}$ and $N_{task}$) and (2) score fusion (fusion between the score upsampling block and the softmax layer).

\renewcommand{\arraystretch}{1.1}
\begin{table}[ht]
	\vspace{-0mm}
	\caption{\textbf{Fusion location}. An evaluation of the impact of network fusion location on final accuracy values. Model: Accel-18. Results for keyframe interval $i=5$, at the max offset ($4$) from the keyframe. Cityscapes dataset.}
	\vspace{3mm}
	\centering
	\label{tbl:fusion}
	\begin{tabular}{@{\extracolsep{4pt}}lcc}
		\toprule
		{} & \multicolumn{2}{c}{Metric} \\
		\cmidrule{2-3}
		Location & Acc. (mIoU) & Time (s/frame) \\
		\midrule
		Feature & \textbf{69.5} & 0.46 \\
		Score & 69.2 & \textbf{0.44} \\
		\bottomrule
	\end{tabular}
	\vspace{-2mm}
\end{table}
\renewcommand{\arraystretch}{1.0}

As Table \ref{tbl:fusion} indicates, score (late) fusion results in slightly lower accuracy, but faster inference times. Recall that a $1 \times 1$ convolutional fusion layer is a mapping $R^{1 \times 2C \times h \times w} \rightarrow R^{1 \times C \times h \times w}$, where $C$ is the channel dimensionality of the input. Feature (early) fusion results in higher accuracy ostensibly because it is executed on higher-dimensionality inputs, allowing for the discovery of richer channel correspondences ($C$ is $2048$ for ResNet feature maps, versus $19$ for scores). Inference times, on the other hand, benefit from \textit{lower} channel dimensionality: the fusion operator itself is cheaper to execute on scores as opposed to features. We use score fusion in all except the most accurate model (Accel-101), as in our view, the 5\% difference in inference cost outweighs the more marginal gap in accuracy. Nevertheless, the choice between the two schemes is a close one.

Finally, we also experimented with the intermediate channel dimensionality, $C$. ResNets-50 and -101 traditionally have channel dimension 2048 after the fifth conv block, which is why $C=2048$ was our default choice. In our experiments, we found that using smaller values of $C$, such as 512 or 1024, resulted in poorer segmentation accuracy, without noticeably reducing inference times.

\subsubsection{Qualitative evaluation}

\begin{figure*}[!ht]
	\vspace{-1mm}
	r1\: \includegraphics[width=4.1cm]{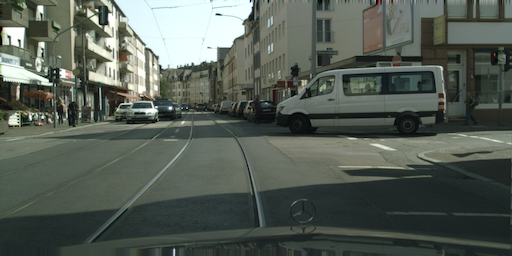} \hfill
	\includegraphics[width=4.1cm]{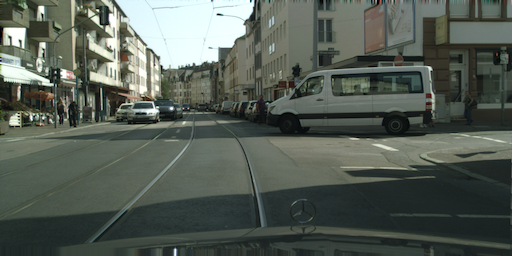} \hfill
	\includegraphics[width=4.1cm]{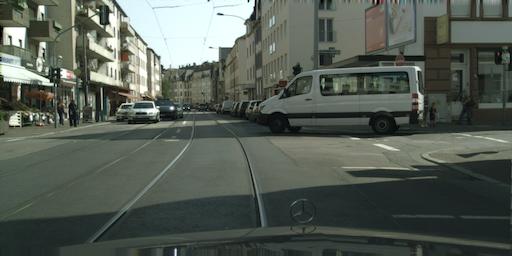} \hfill
	\includegraphics[width=4.1cm]{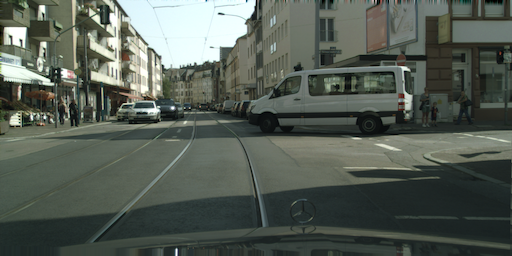} \\[1pt]
	r2\: \includegraphics[width=4.1cm]{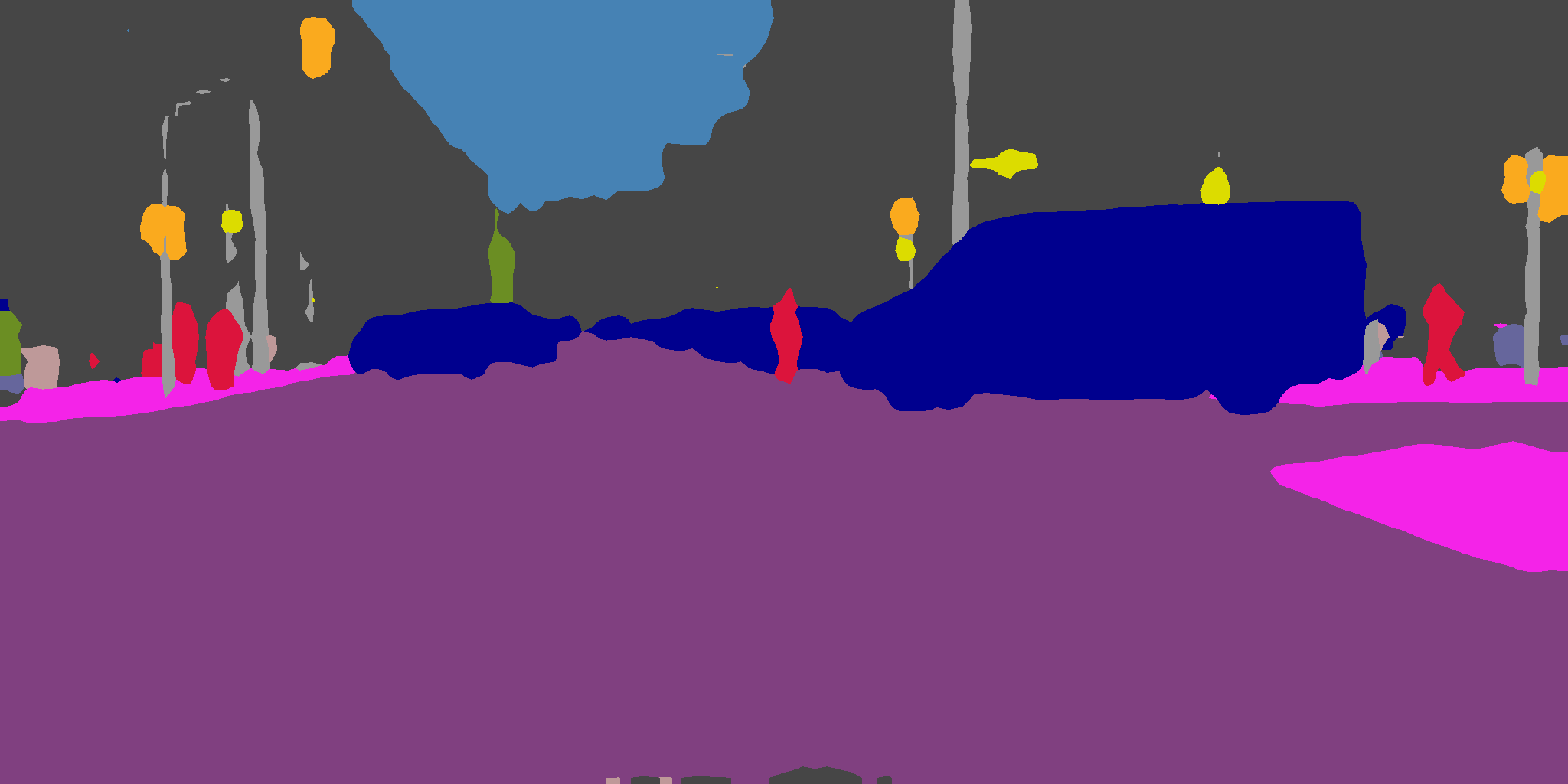} \hfill
	\includegraphics[width=4.1cm]{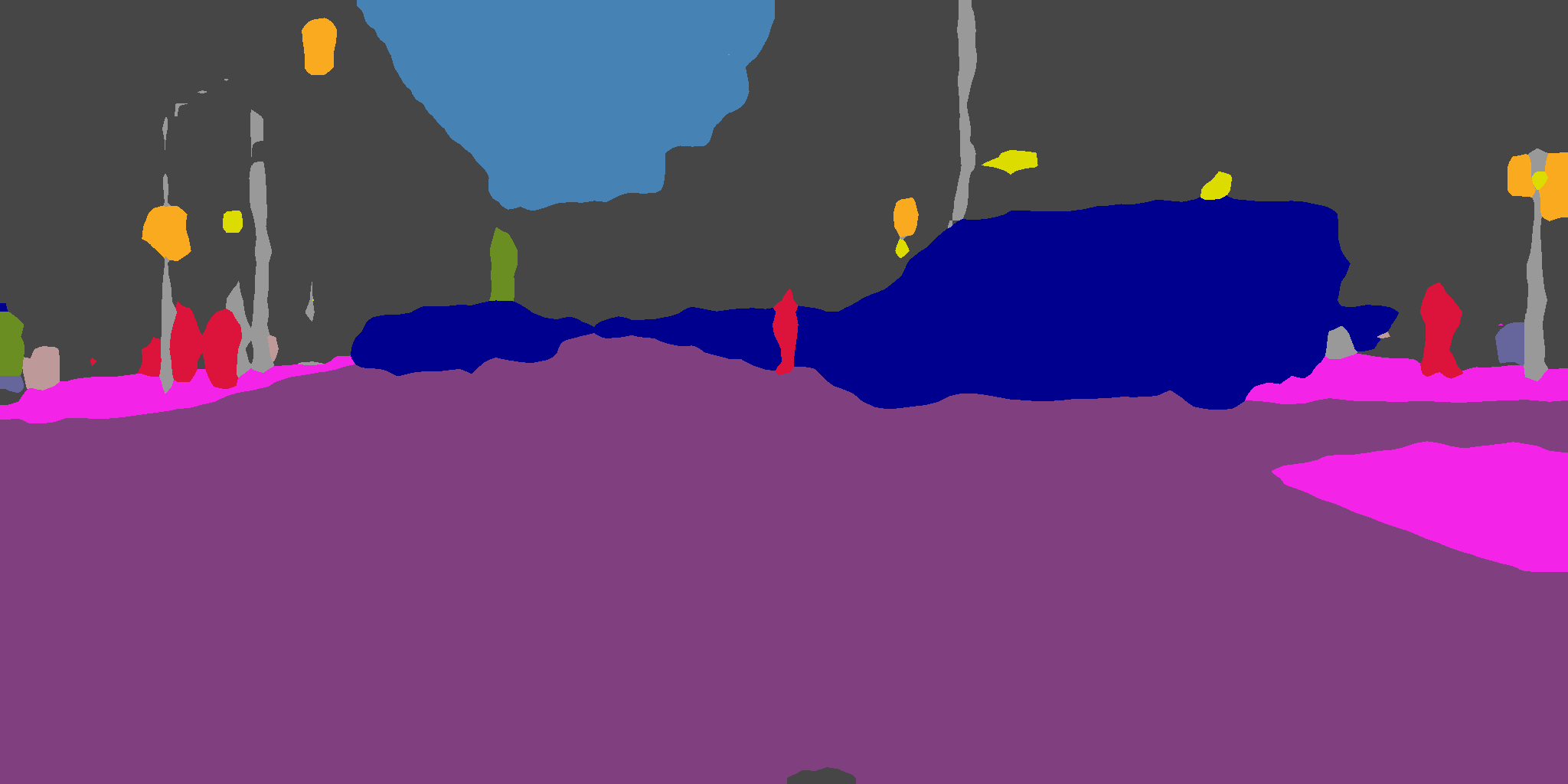} \hfill
	\includegraphics[width=4.1cm]{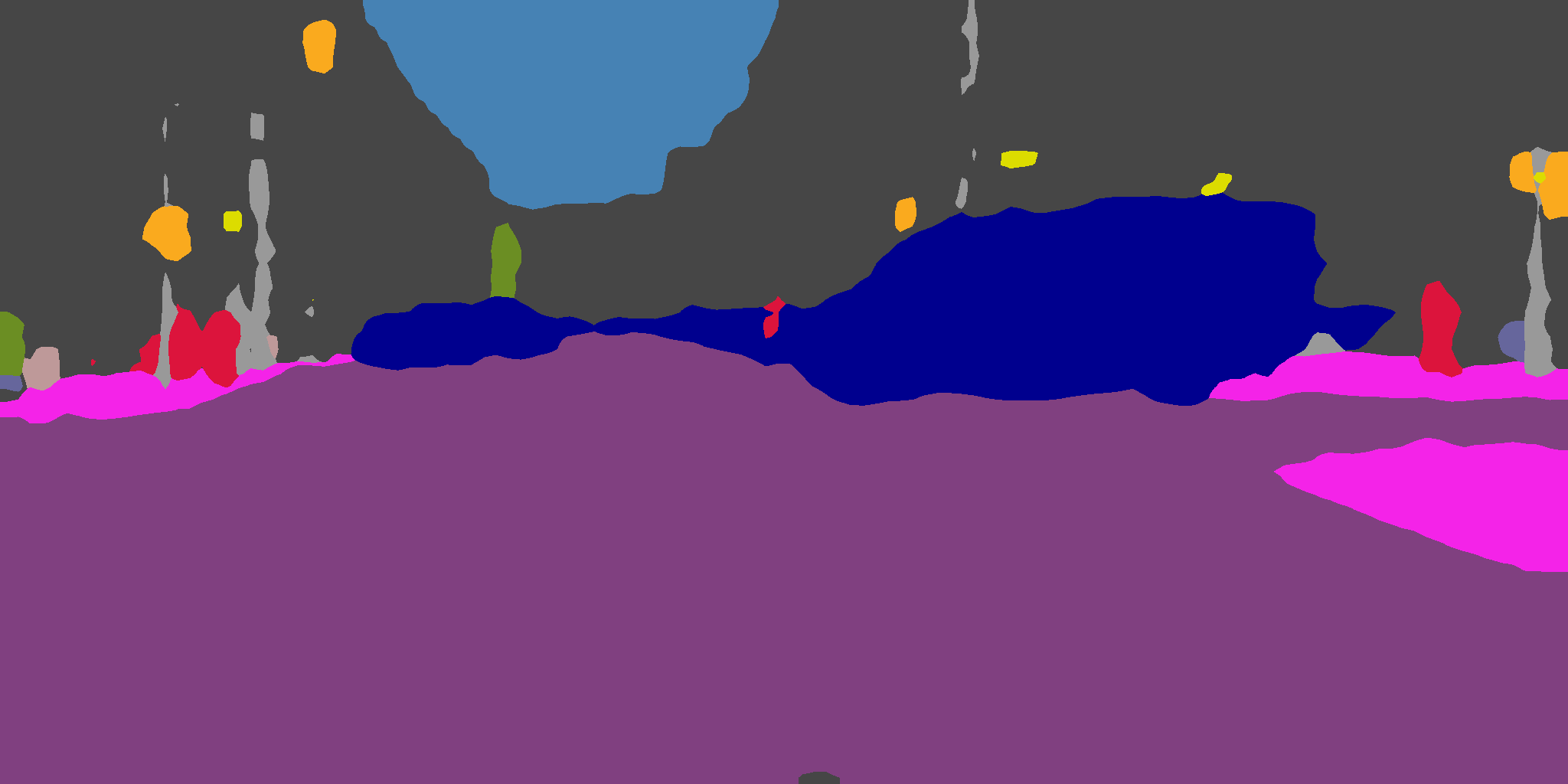} \hfill
	\includegraphics[width=4.1cm]{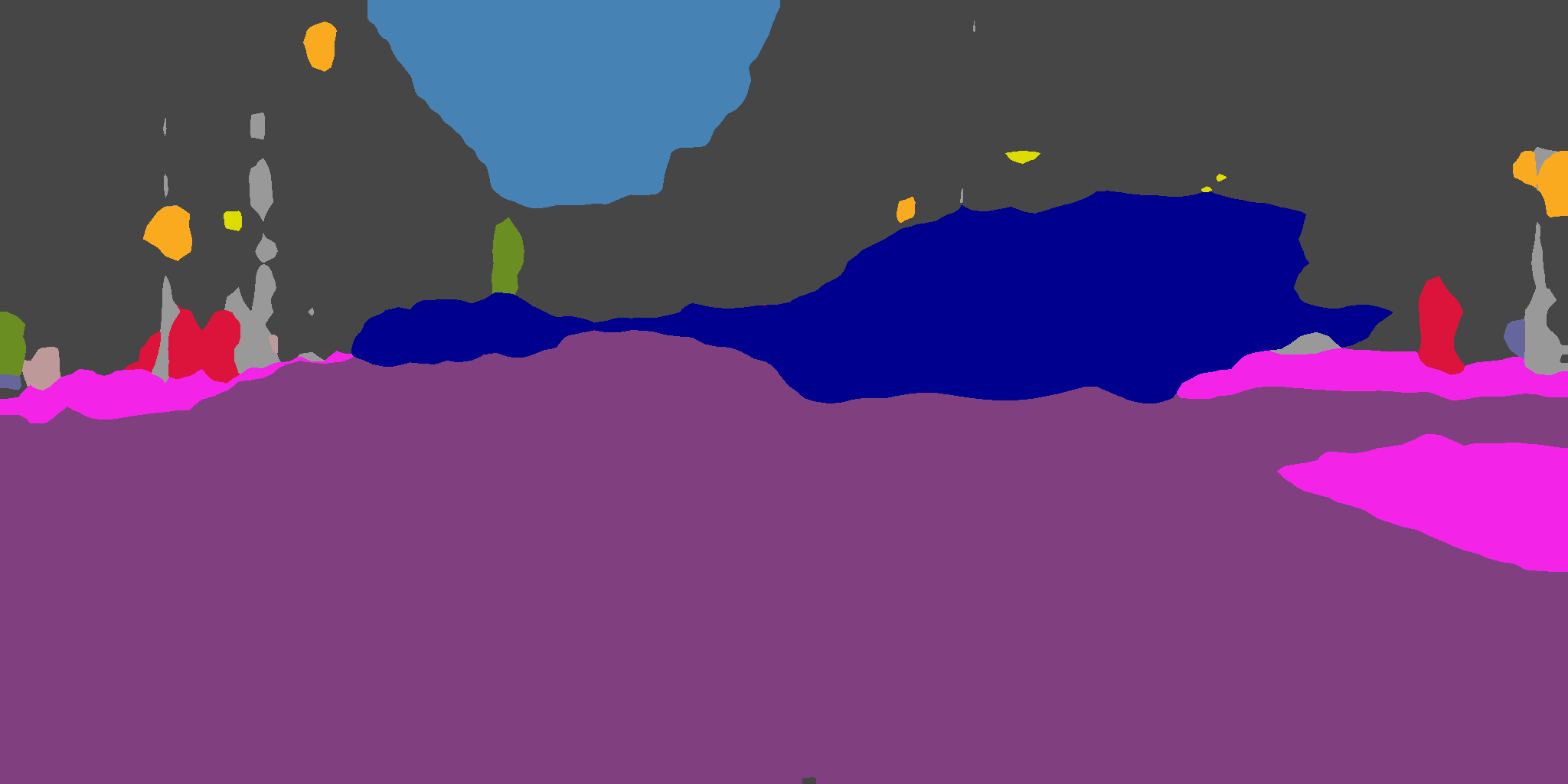} \\[1pt]
	r3\: \includegraphics[width=4.1cm]{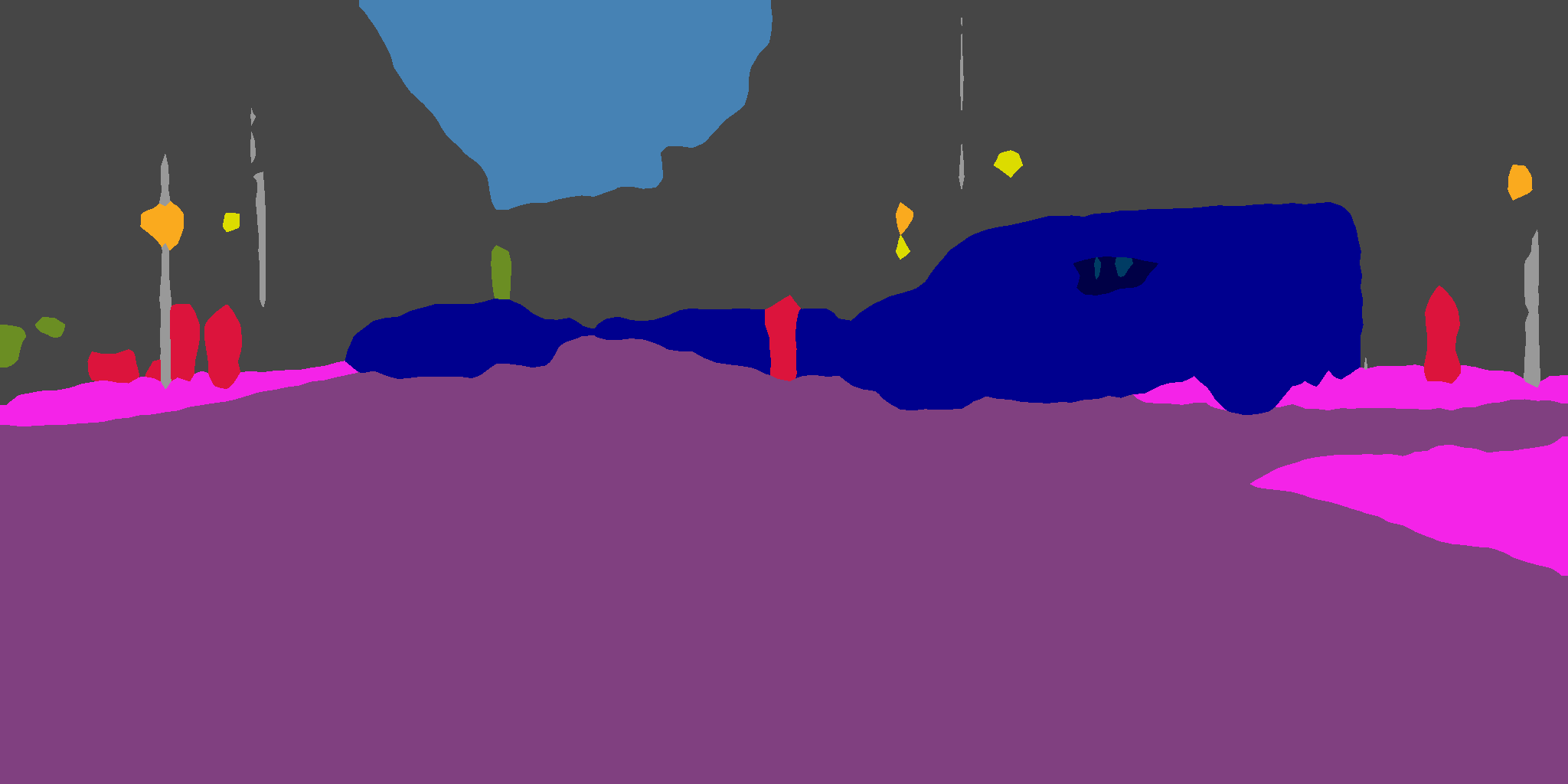} \hfill
	\includegraphics[width=4.1cm]{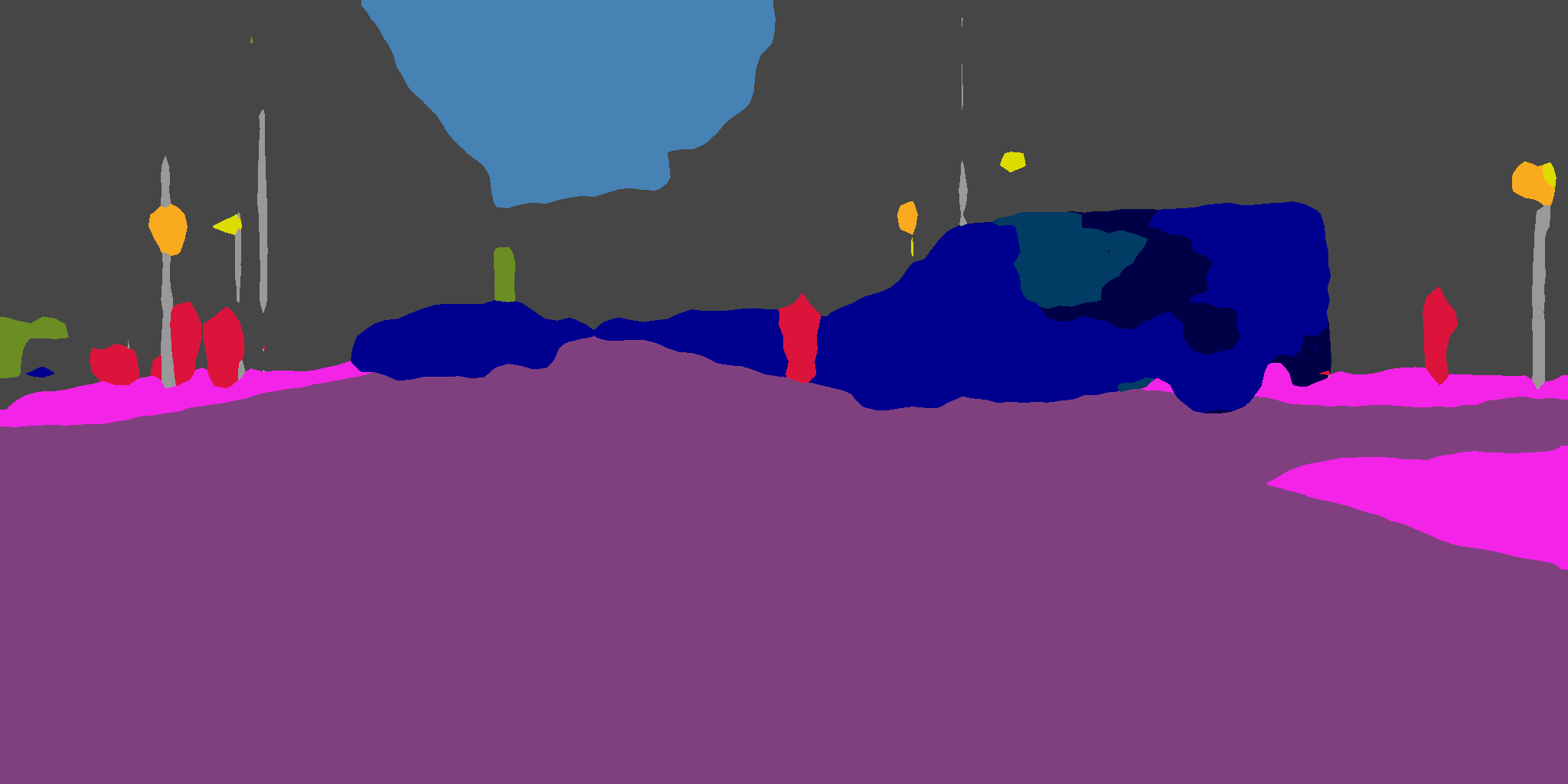} \hfill
	\includegraphics[width=4.1cm]{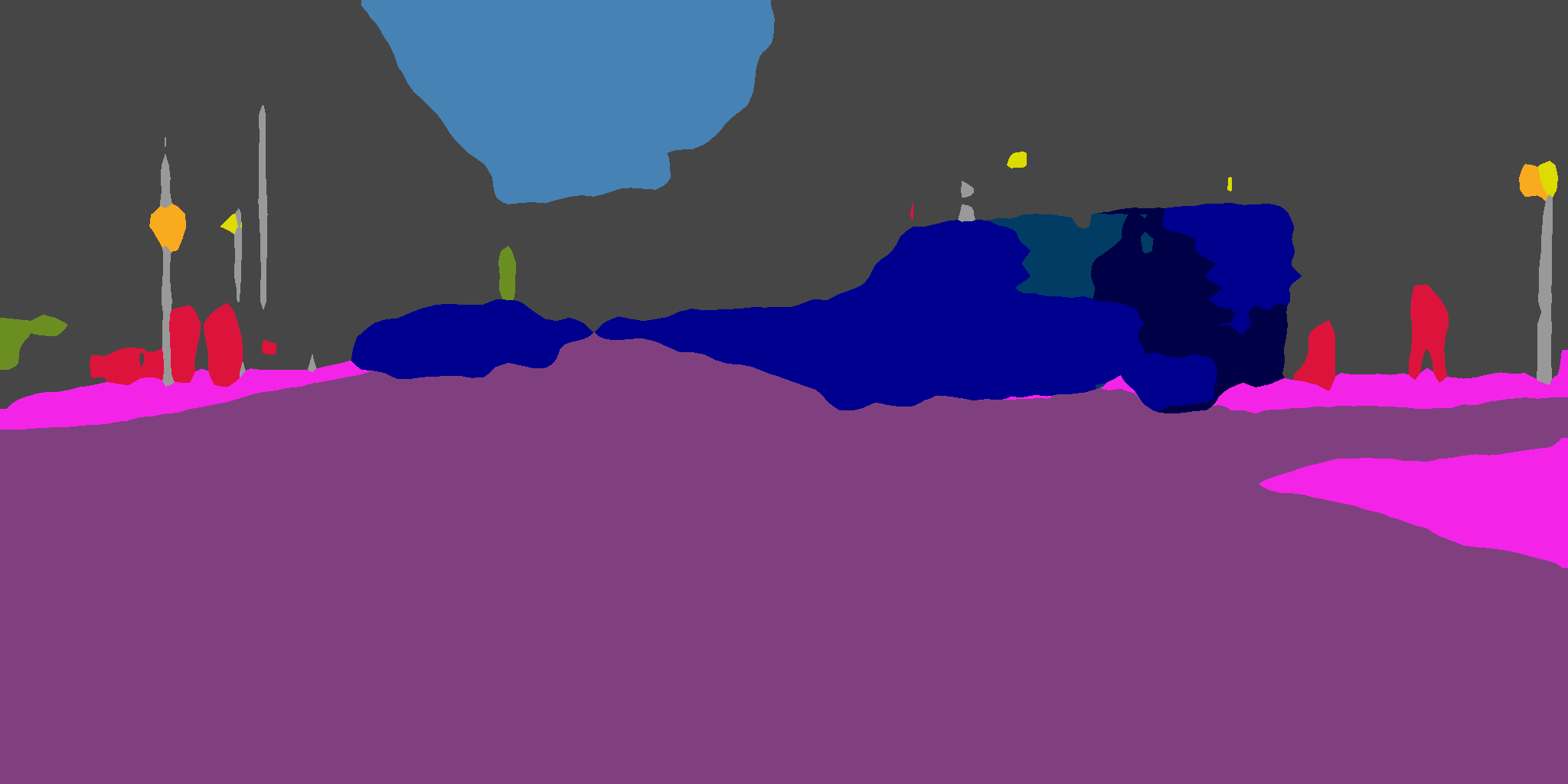} \hfill
	\includegraphics[width=4.1cm]{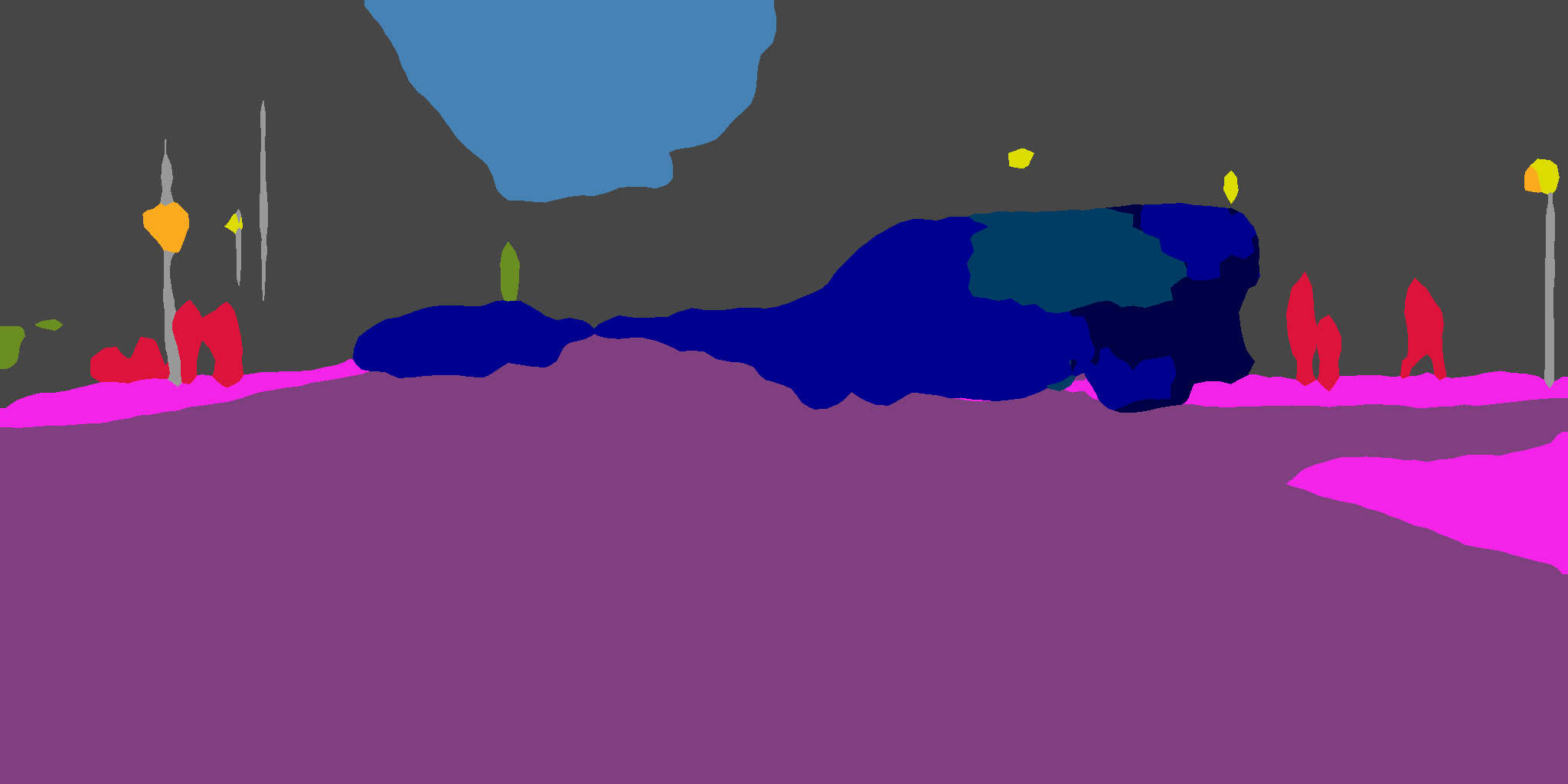} \\[1pt]
	r4\: \includegraphics[width=4.1cm]{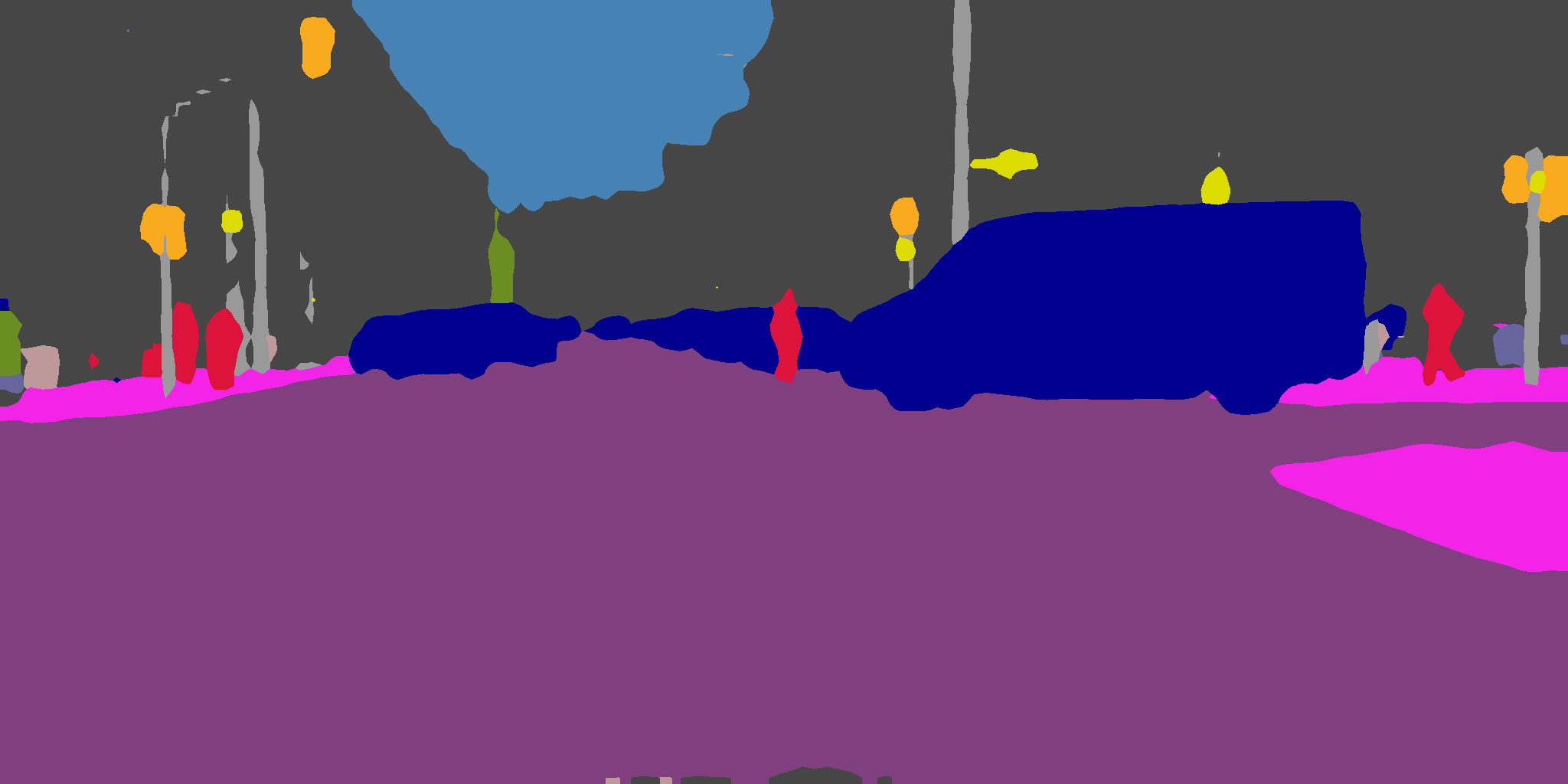} \hfill
	\includegraphics[width=4.1cm]{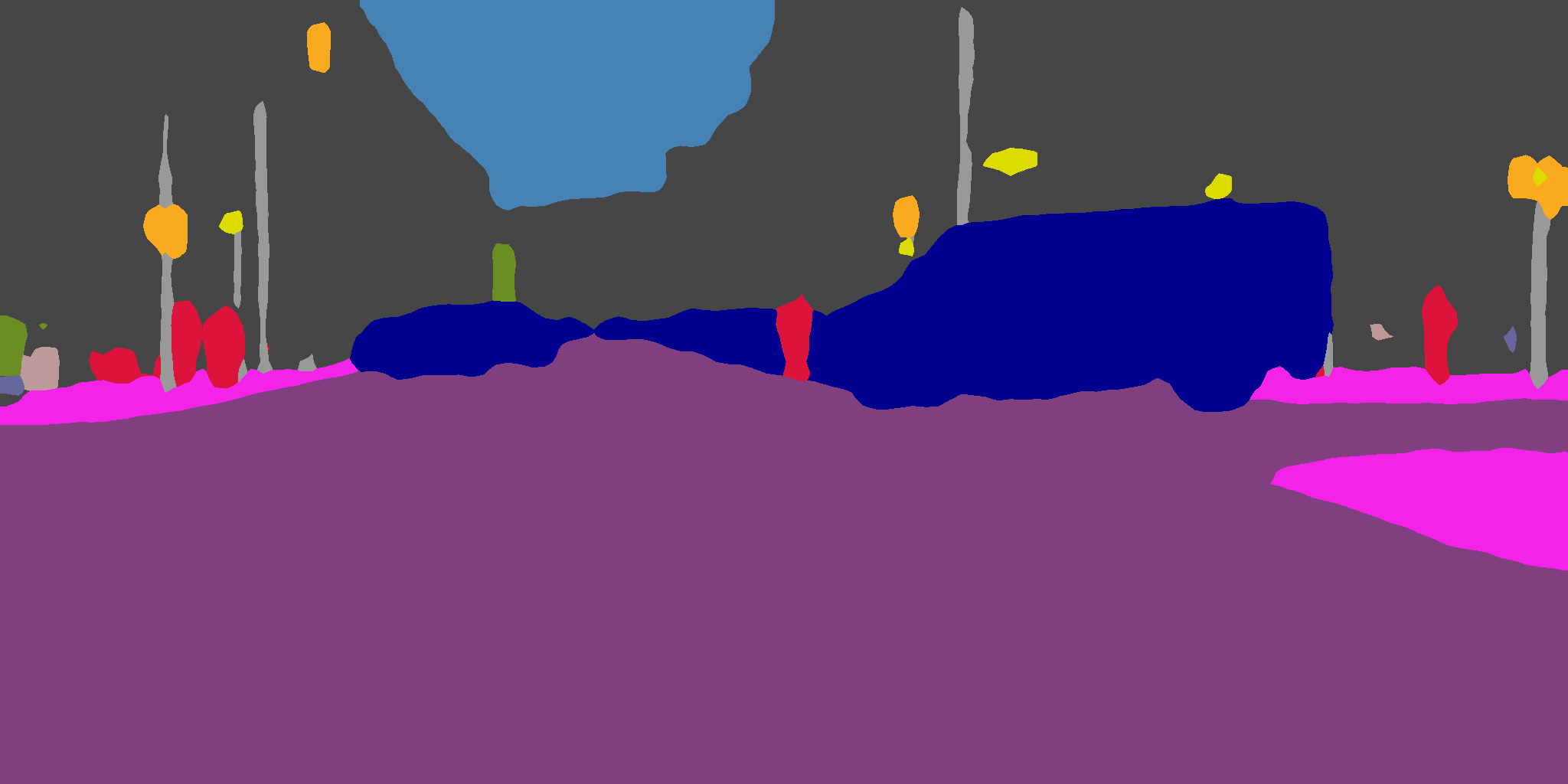} \hfill
	\includegraphics[width=4.1cm]{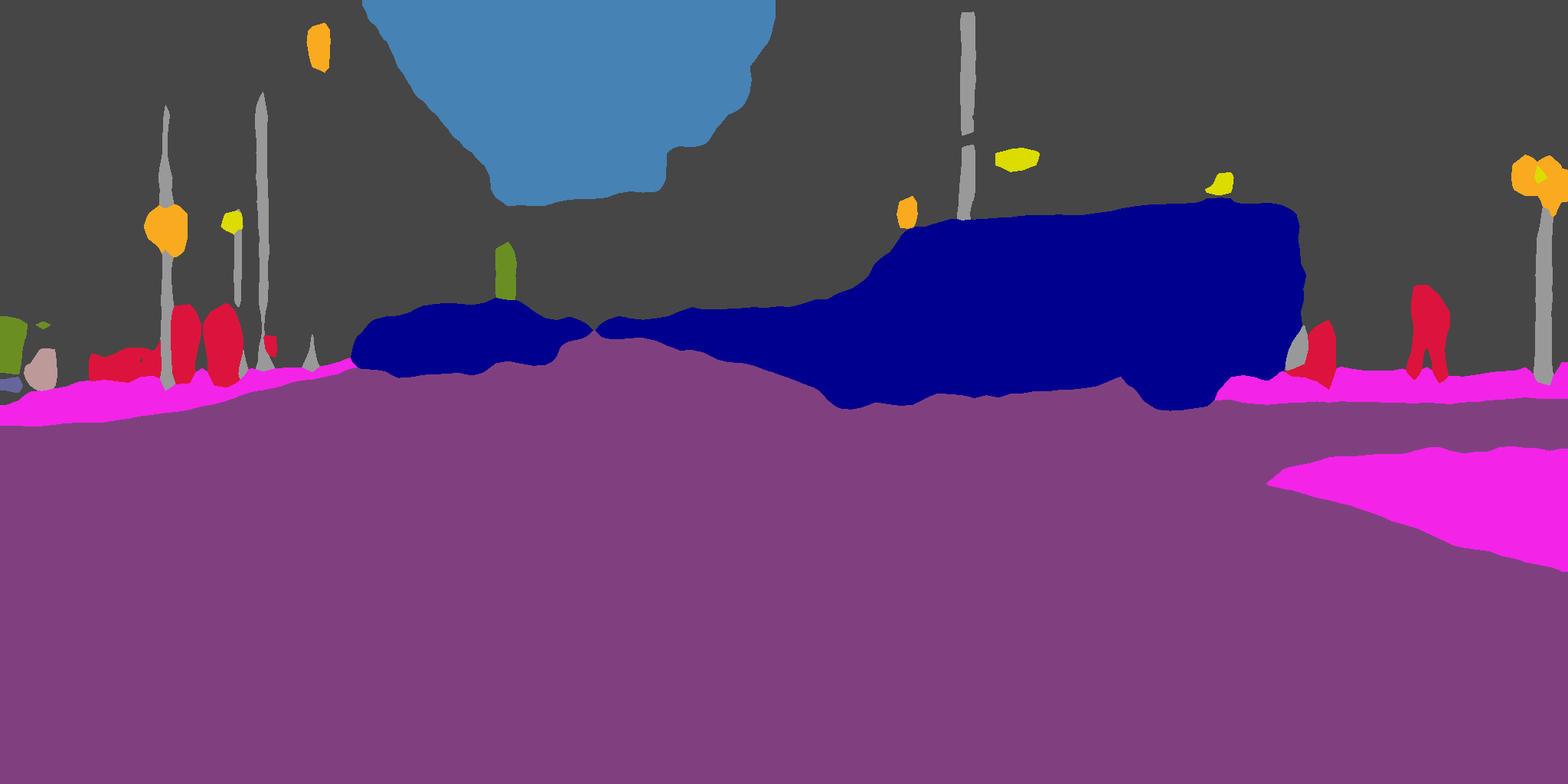} \hfill
	\includegraphics[width=4.1cm]{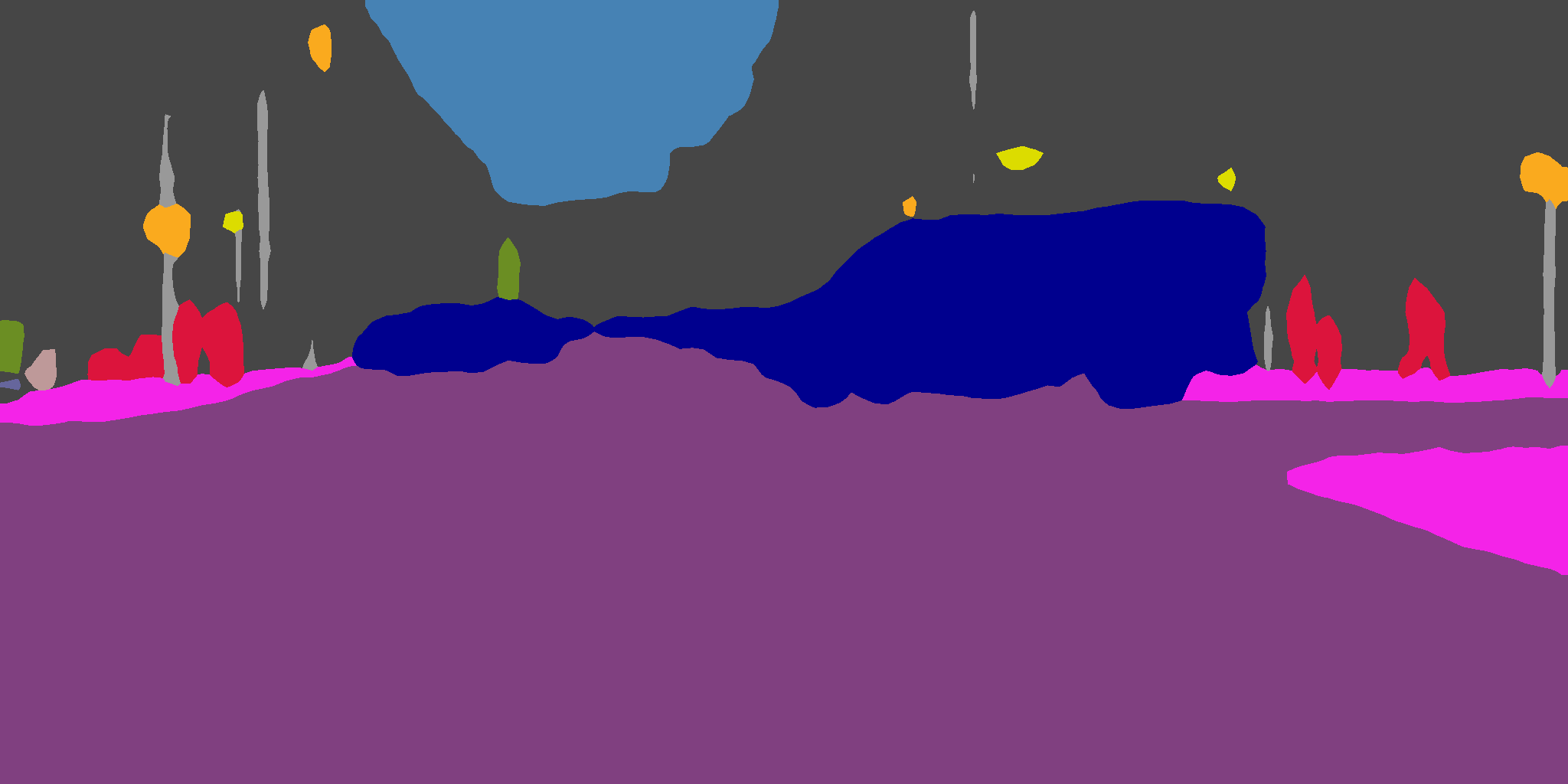} \\[2pt]
	
	r1\: \includegraphics[width=4.1cm]{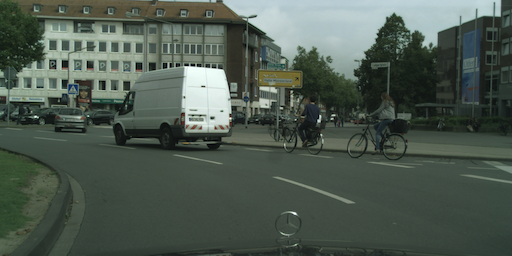} \hfill
	\includegraphics[width=4.1cm]{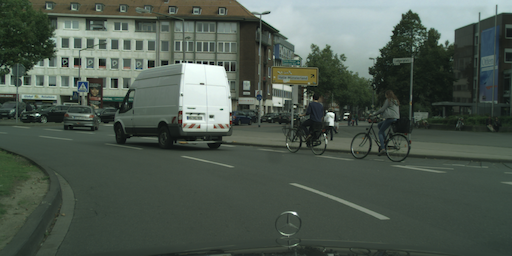} \hfill
	\includegraphics[width=4.1cm]{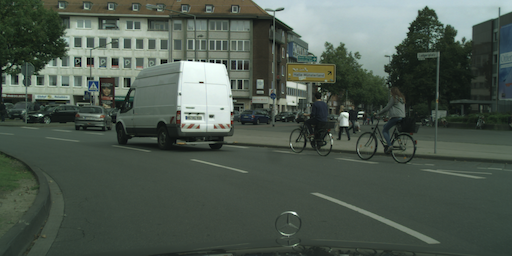} \hfill
	\includegraphics[width=4.1cm]{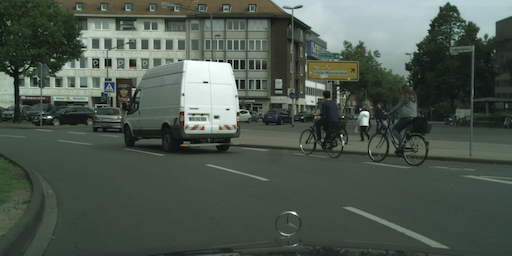} \\[1pt]
	r2\: \includegraphics[width=4.1cm]{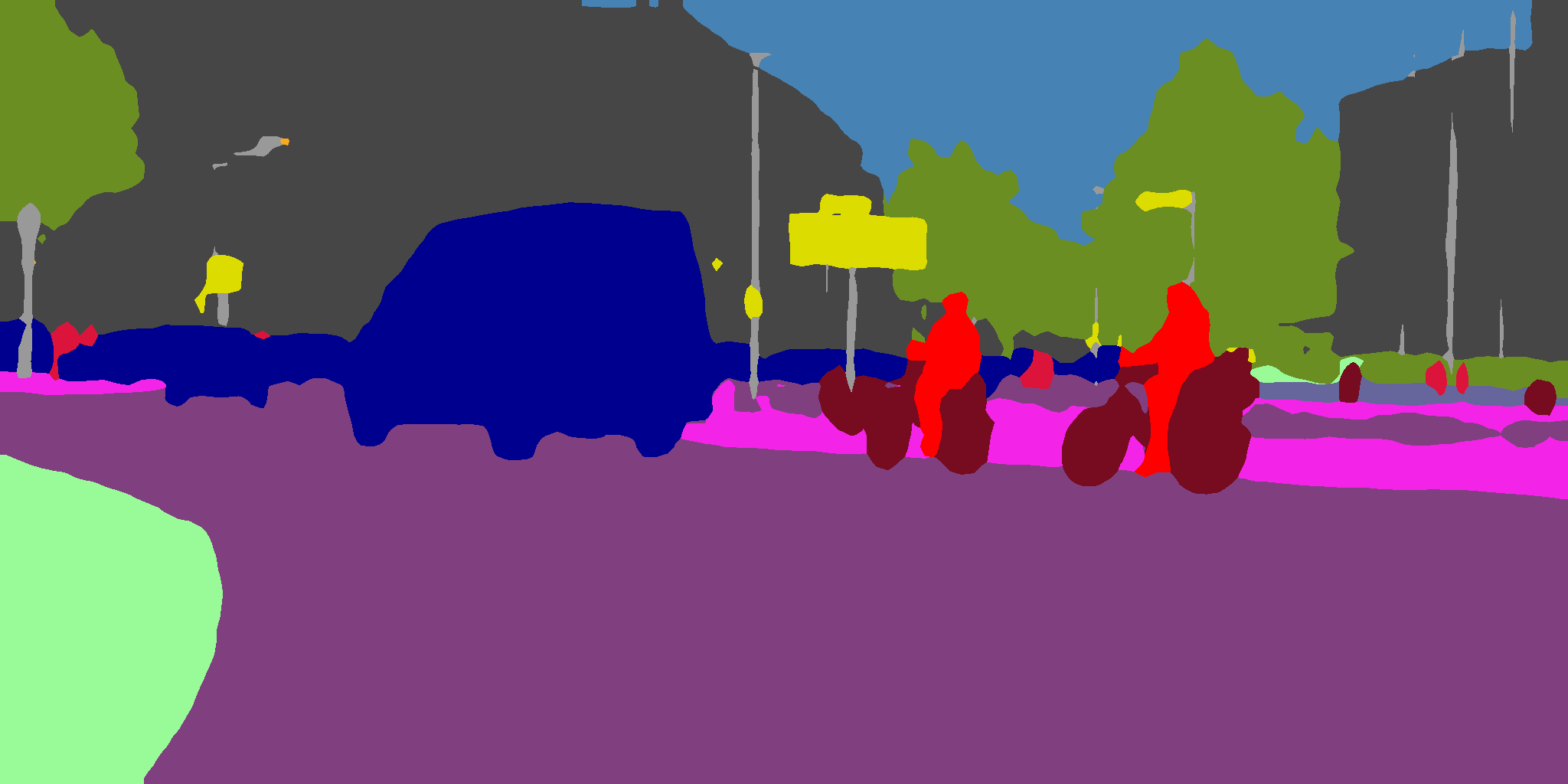} \hfill
	\includegraphics[width=4.1cm]{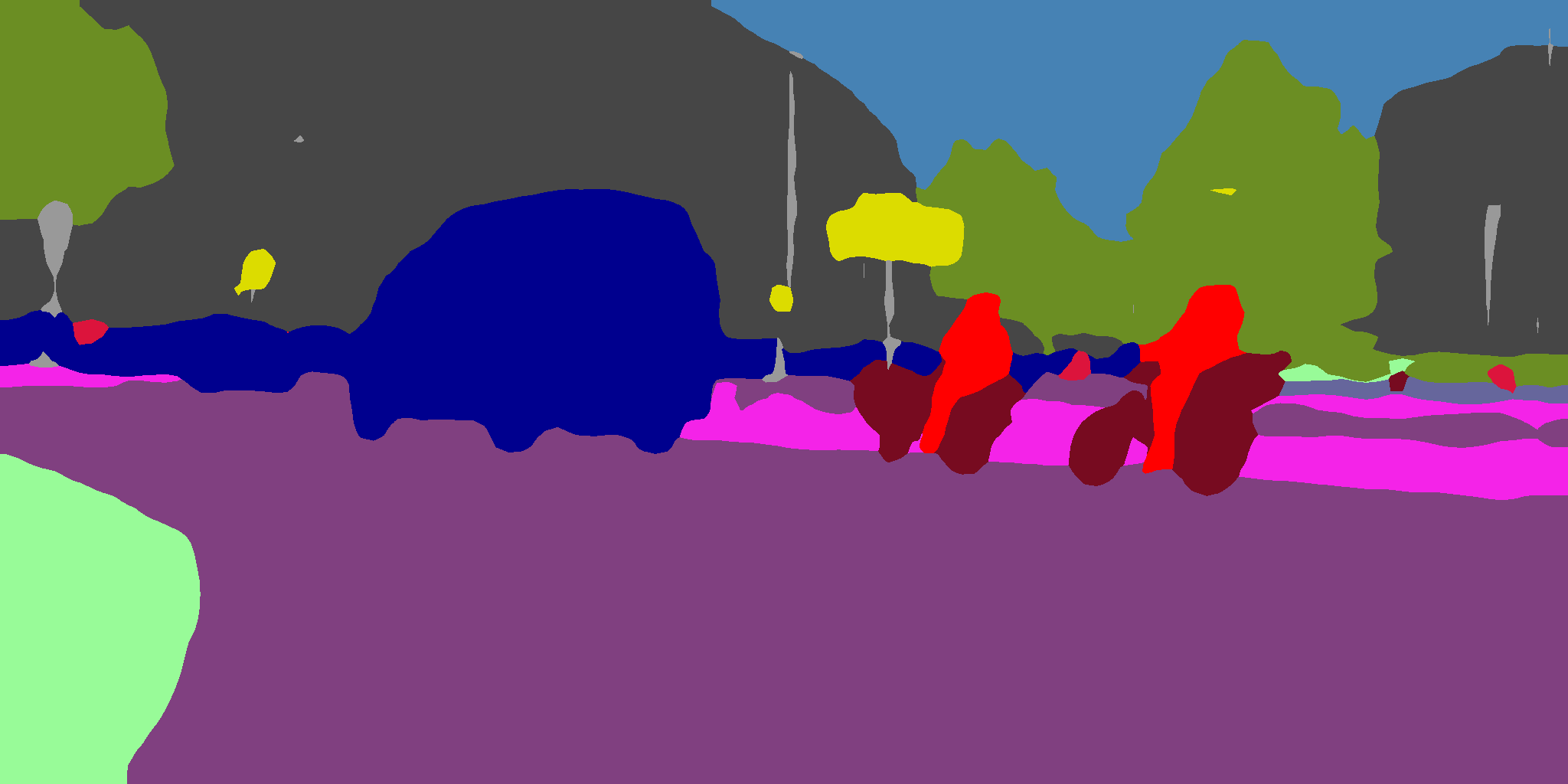} \hfill
	\includegraphics[width=4.1cm]{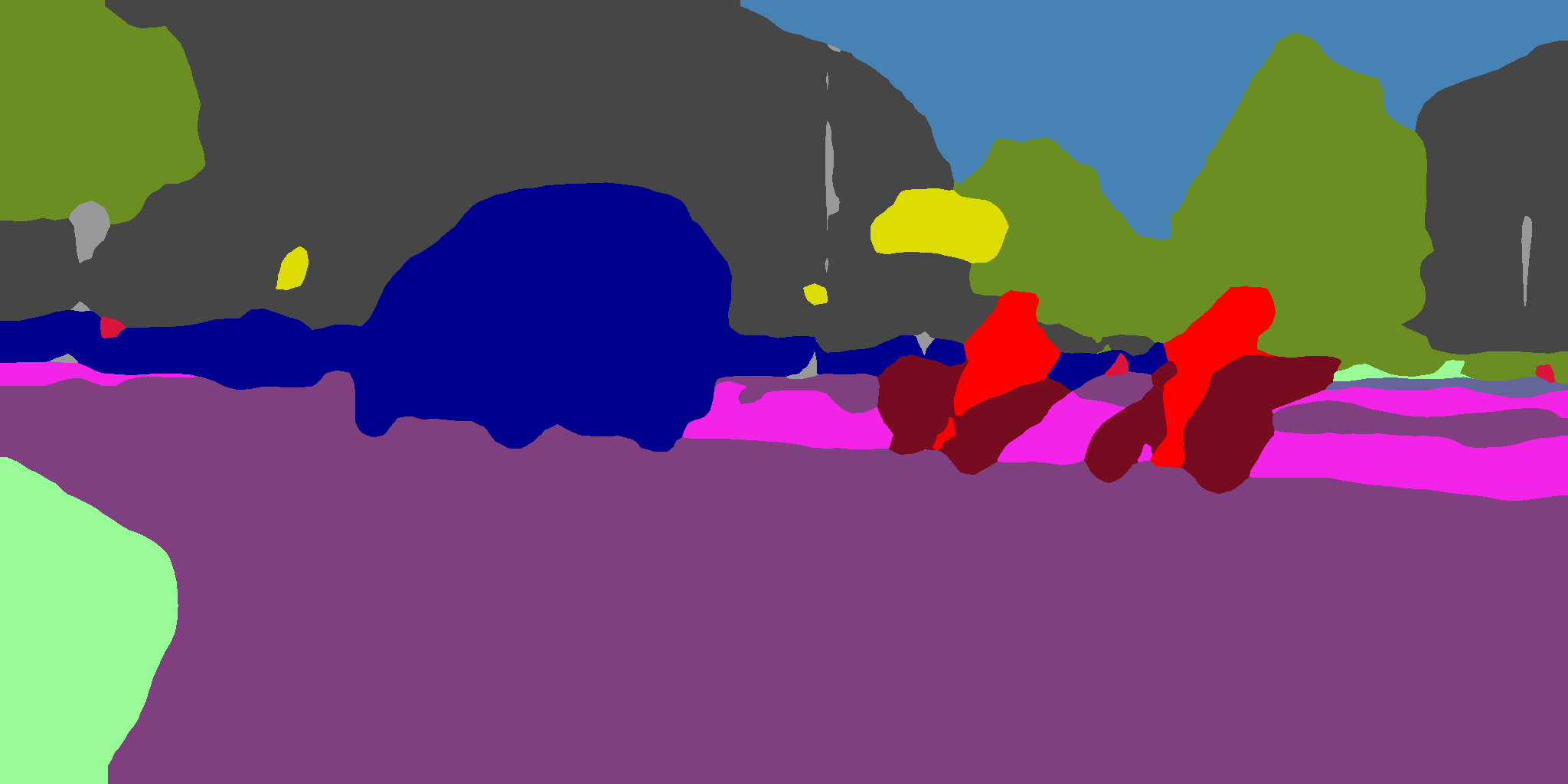} \hfill
	\includegraphics[width=4.1cm]{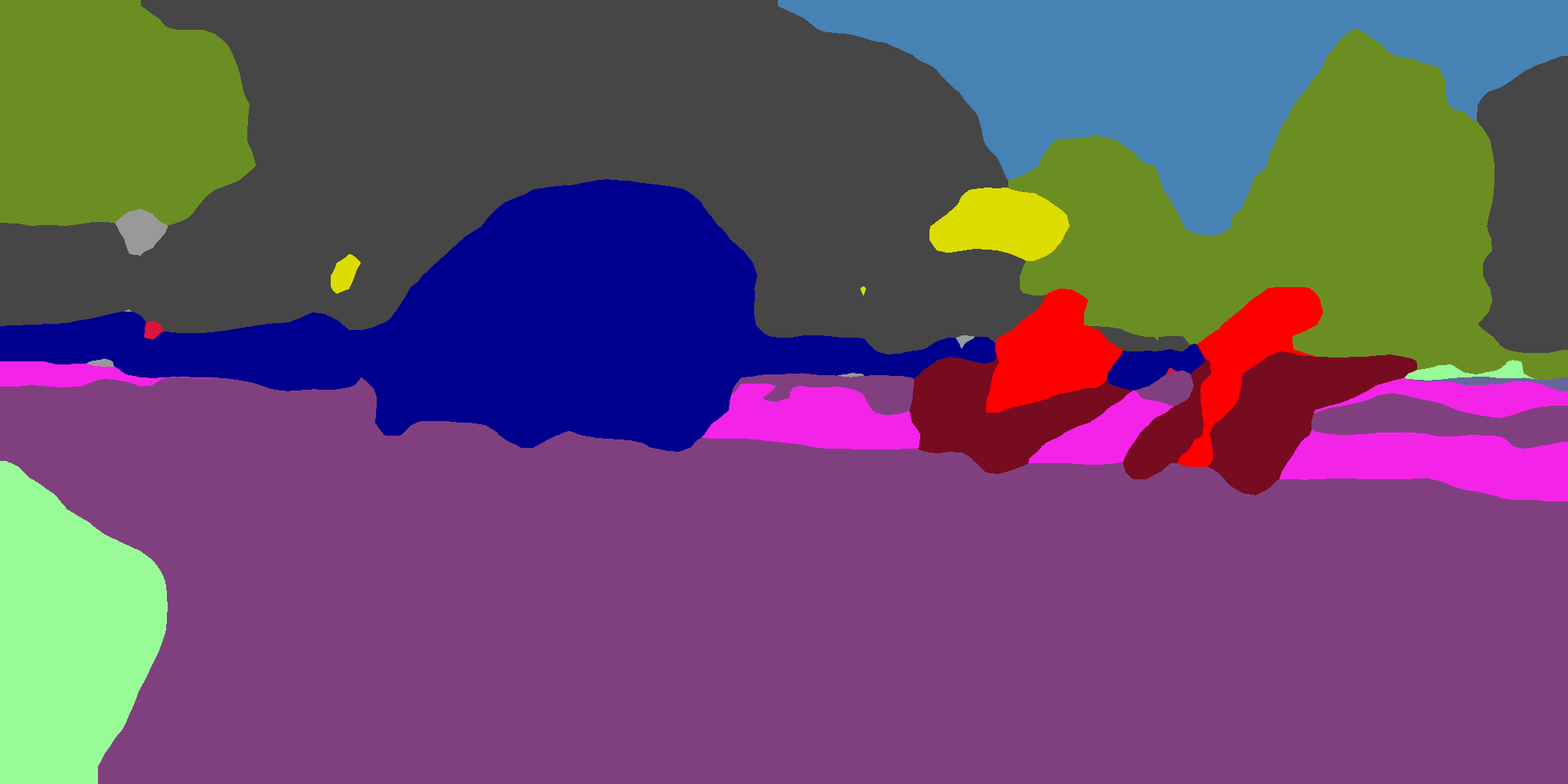} \\[1pt]
	r3\: \includegraphics[width=4.1cm]{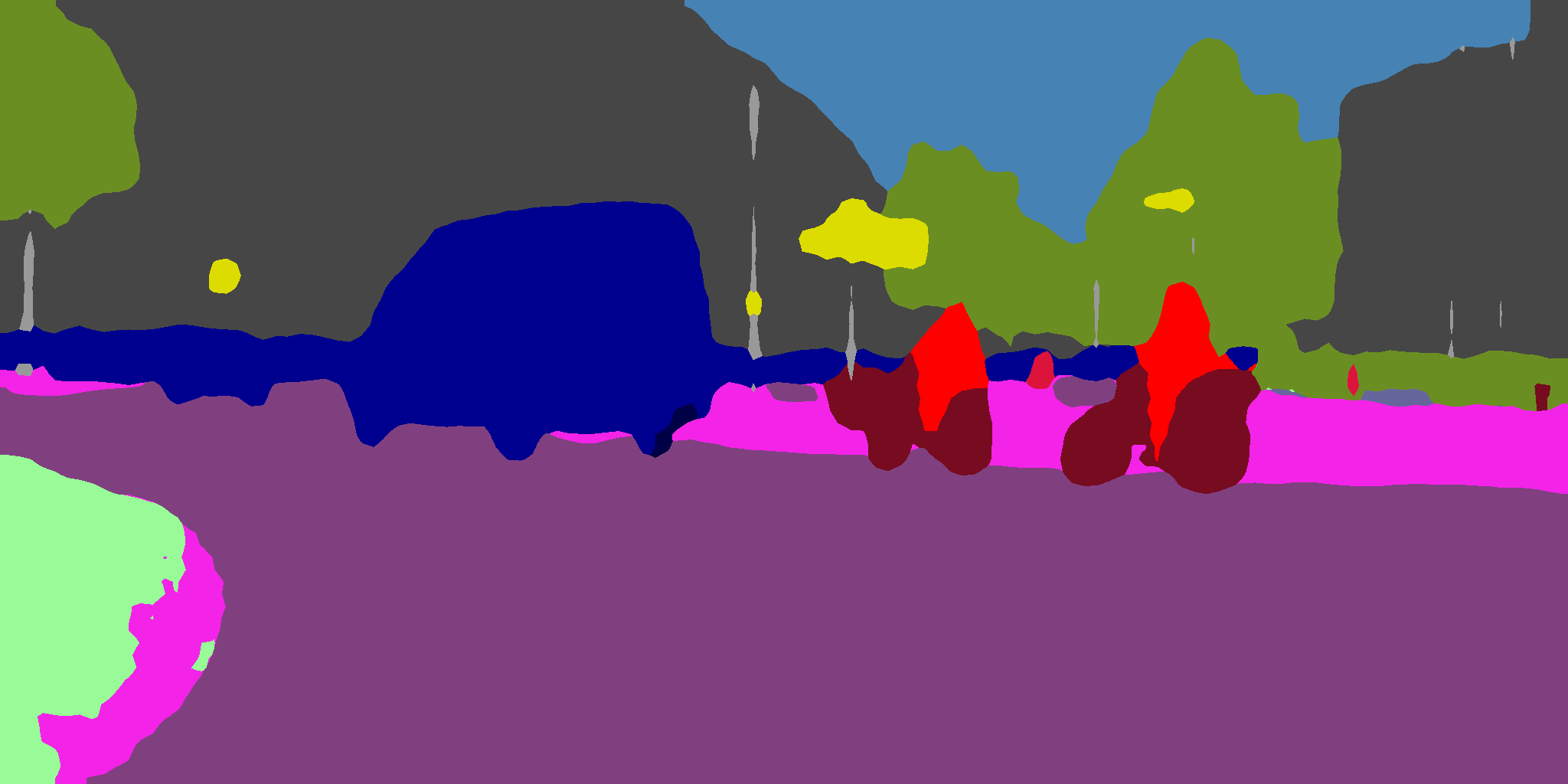} \hfill
	\includegraphics[width=4.1cm]{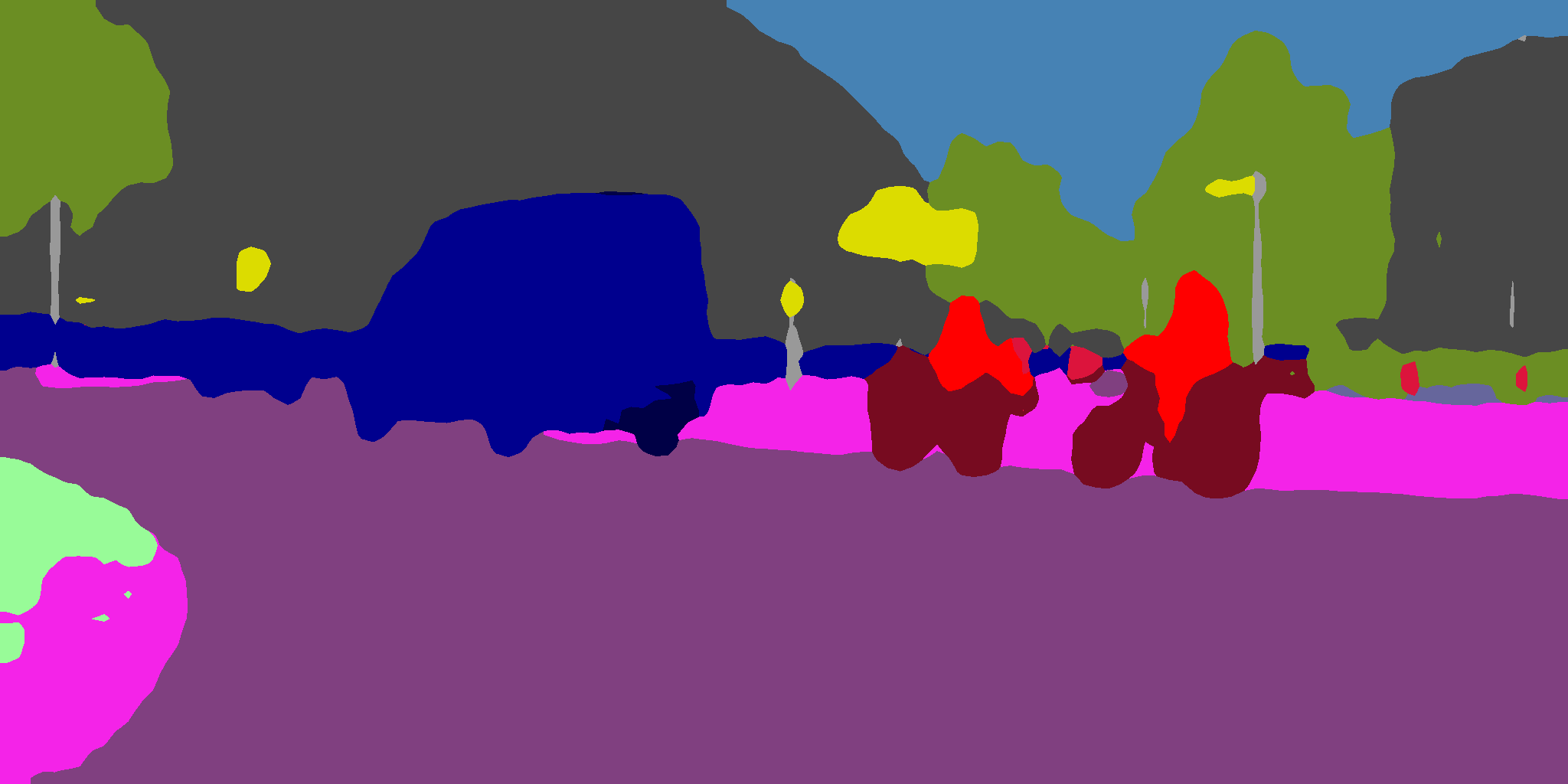} \hfill
	\includegraphics[width=4.1cm]{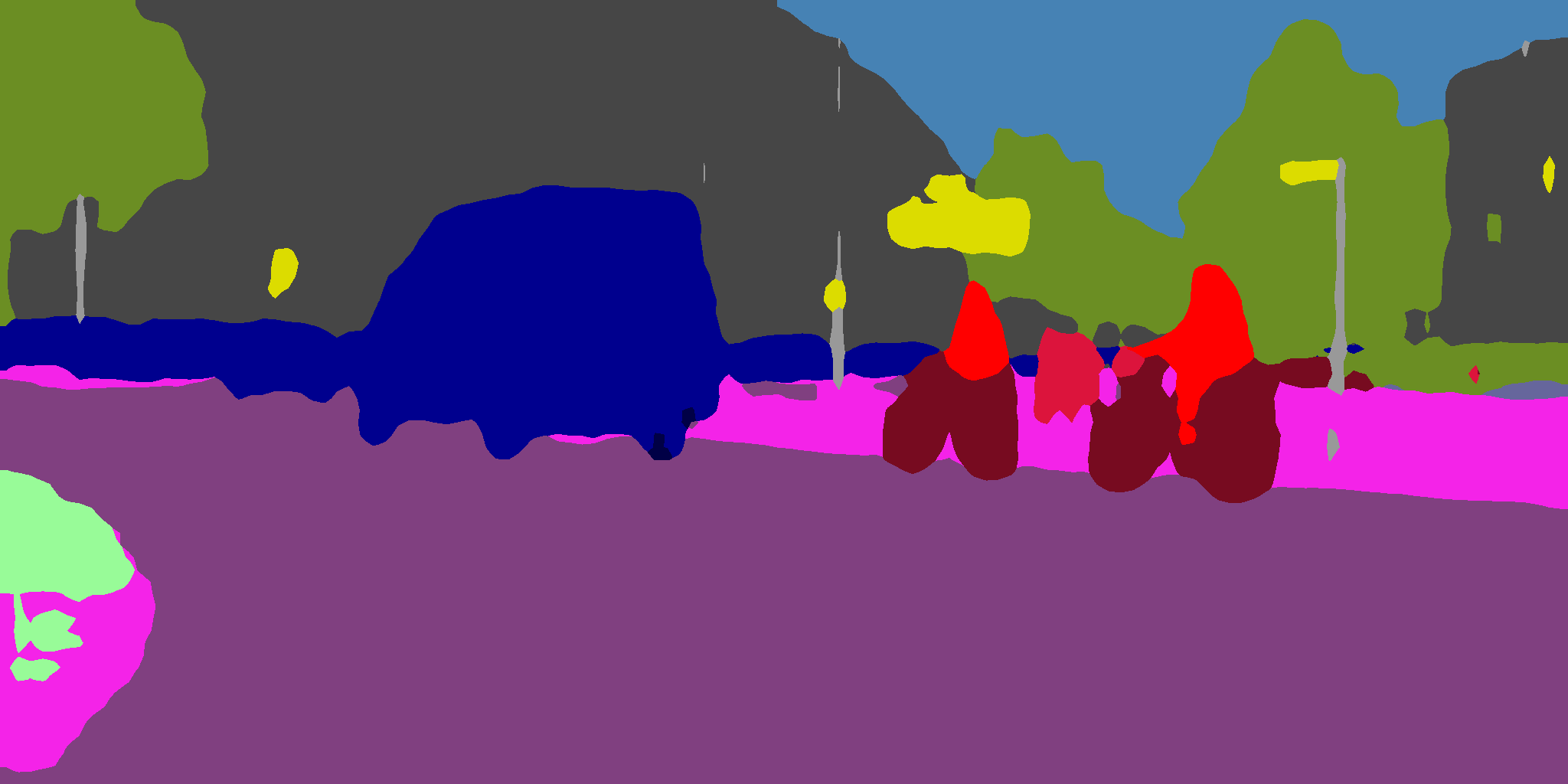} \hfill
	\includegraphics[width=4.1cm]{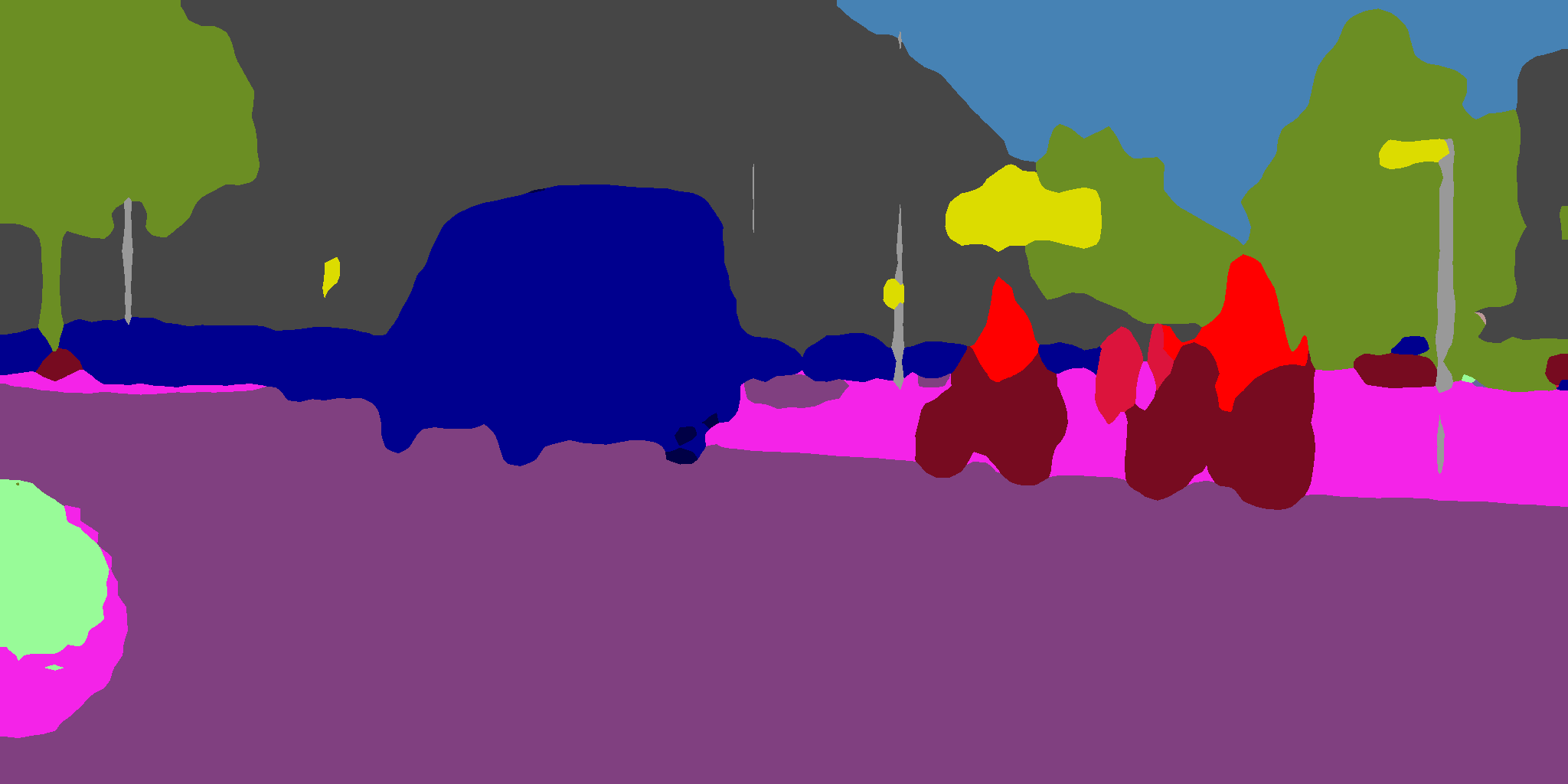} \\[-9pt]
	r4\: \subfloat[$k$    ]{\includegraphics[width=4.1cm]{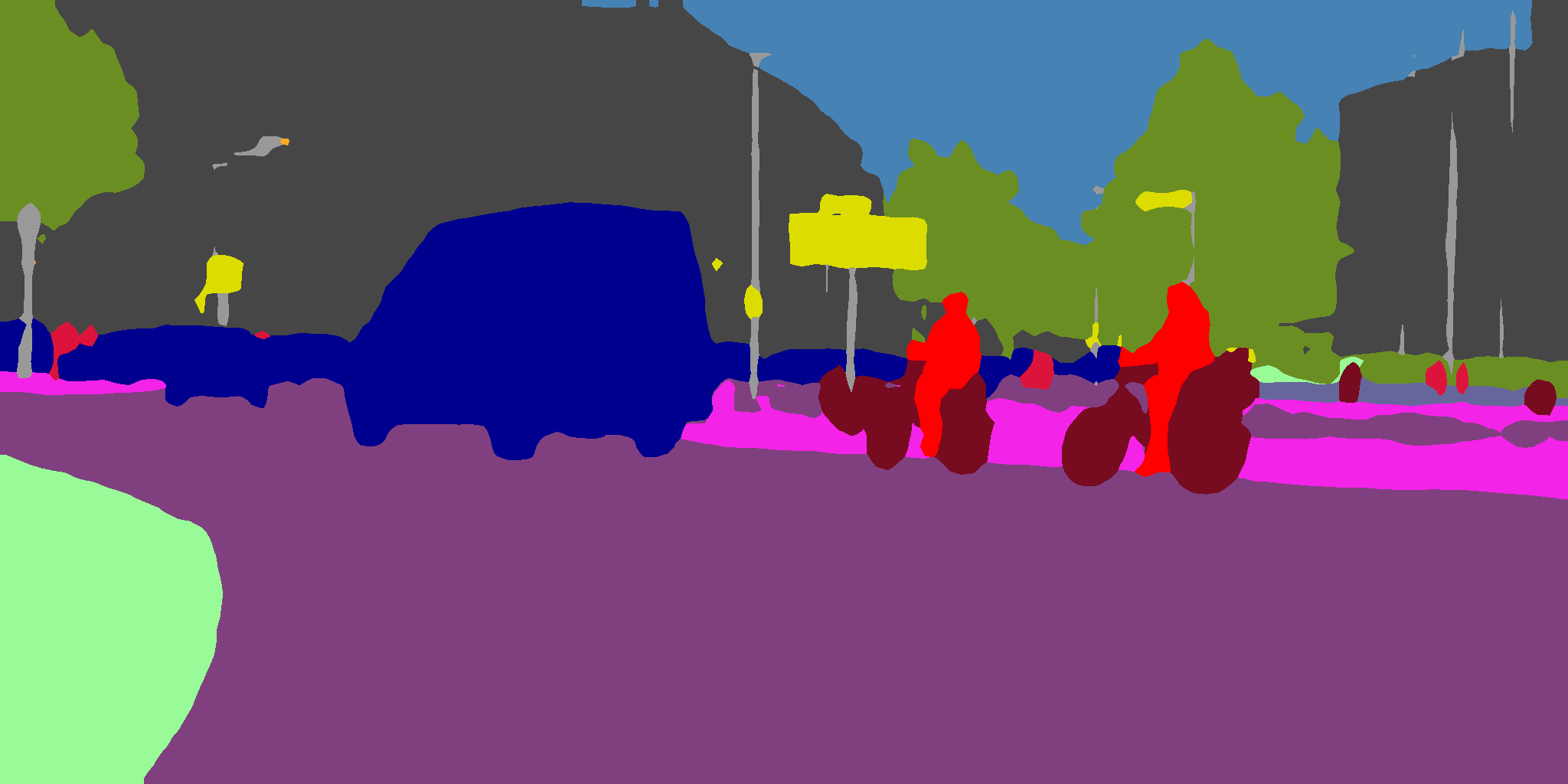}} \hfill
	\subfloat[$k+3$]{\includegraphics[width=4.1cm]{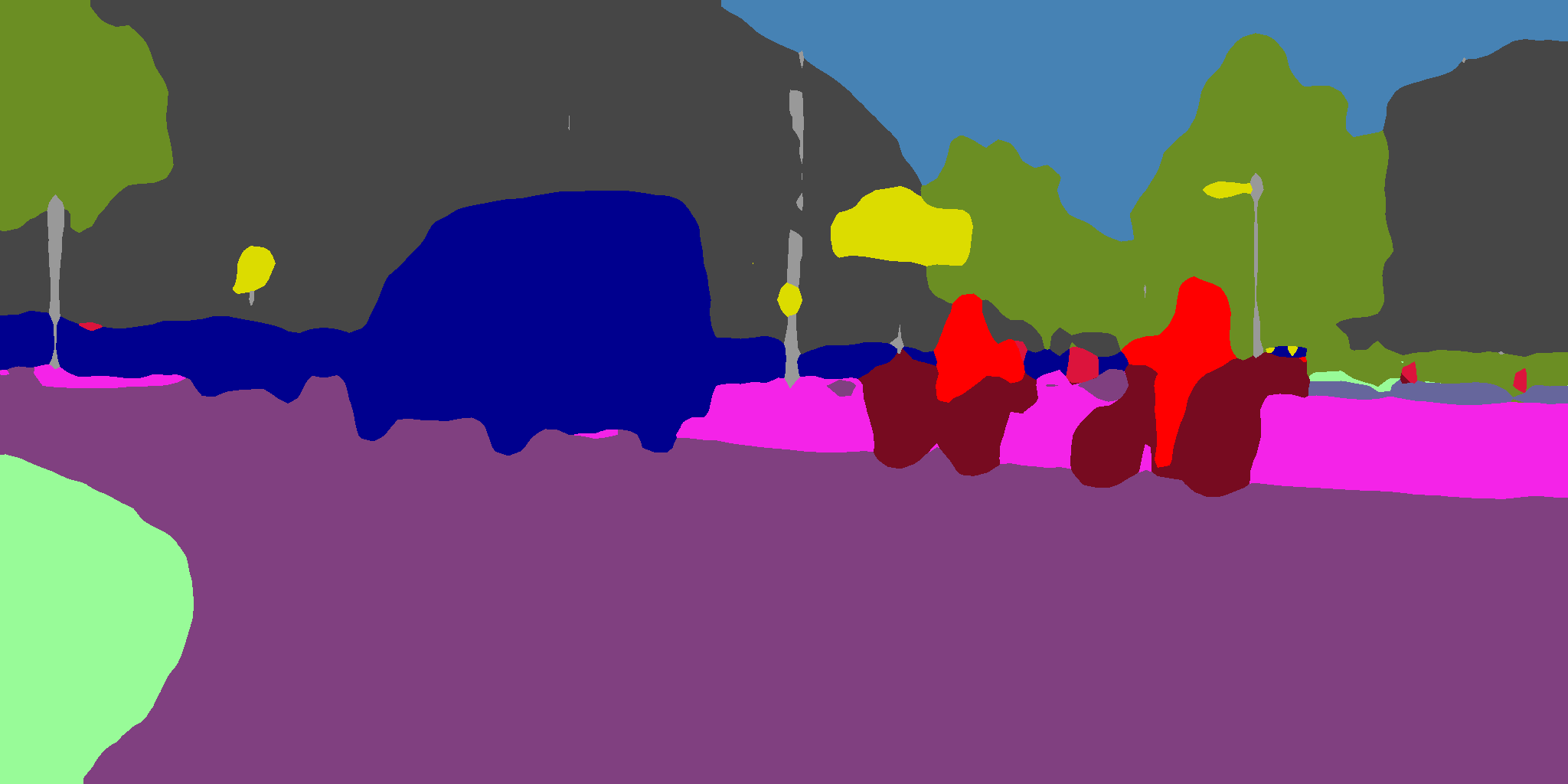}} \hfill
	\subfloat[$k+6$]{\includegraphics[width=4.1cm]{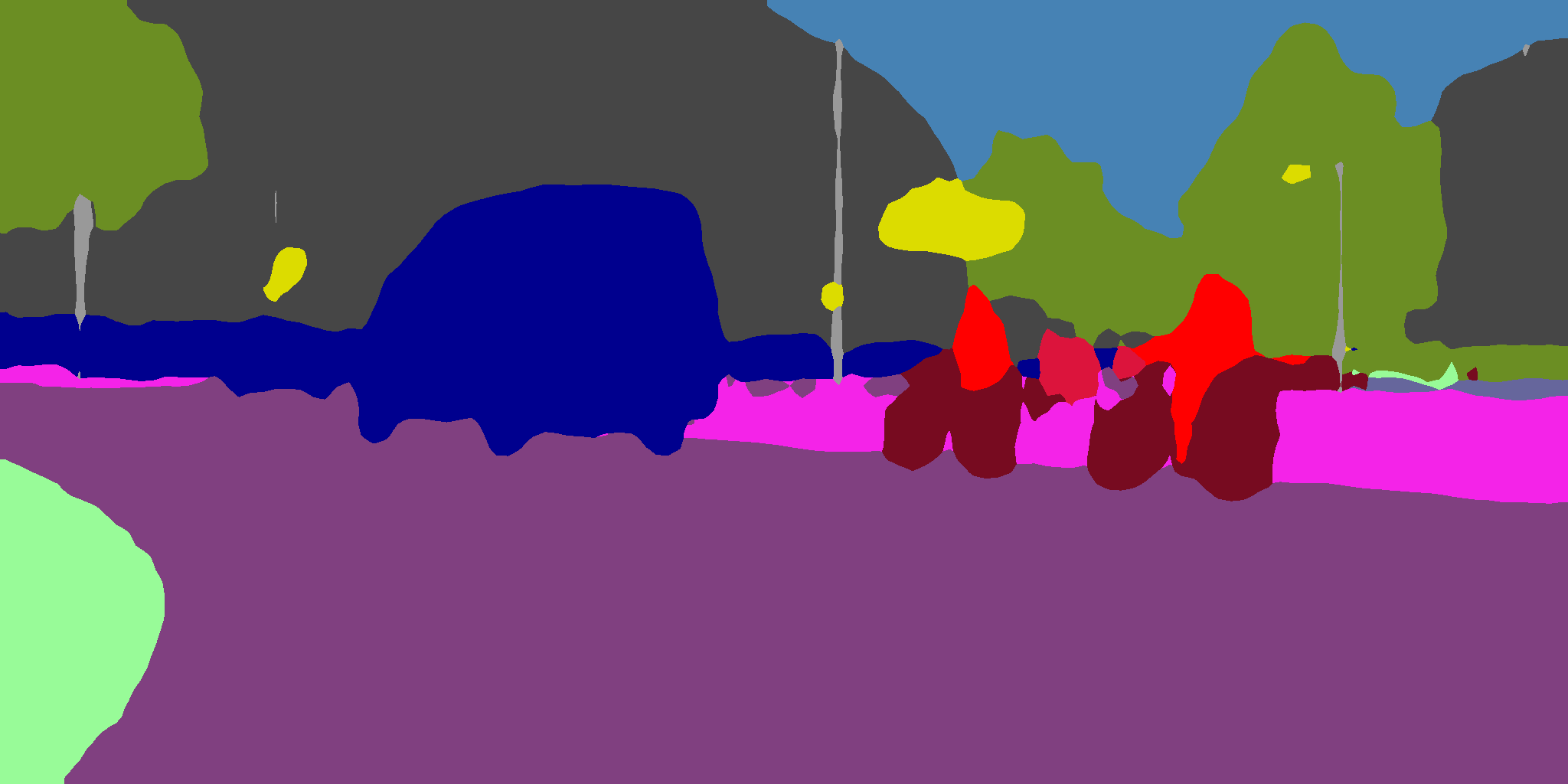}} \hfill
	\subfloat[$k+9$]{\includegraphics[width=4.1cm]{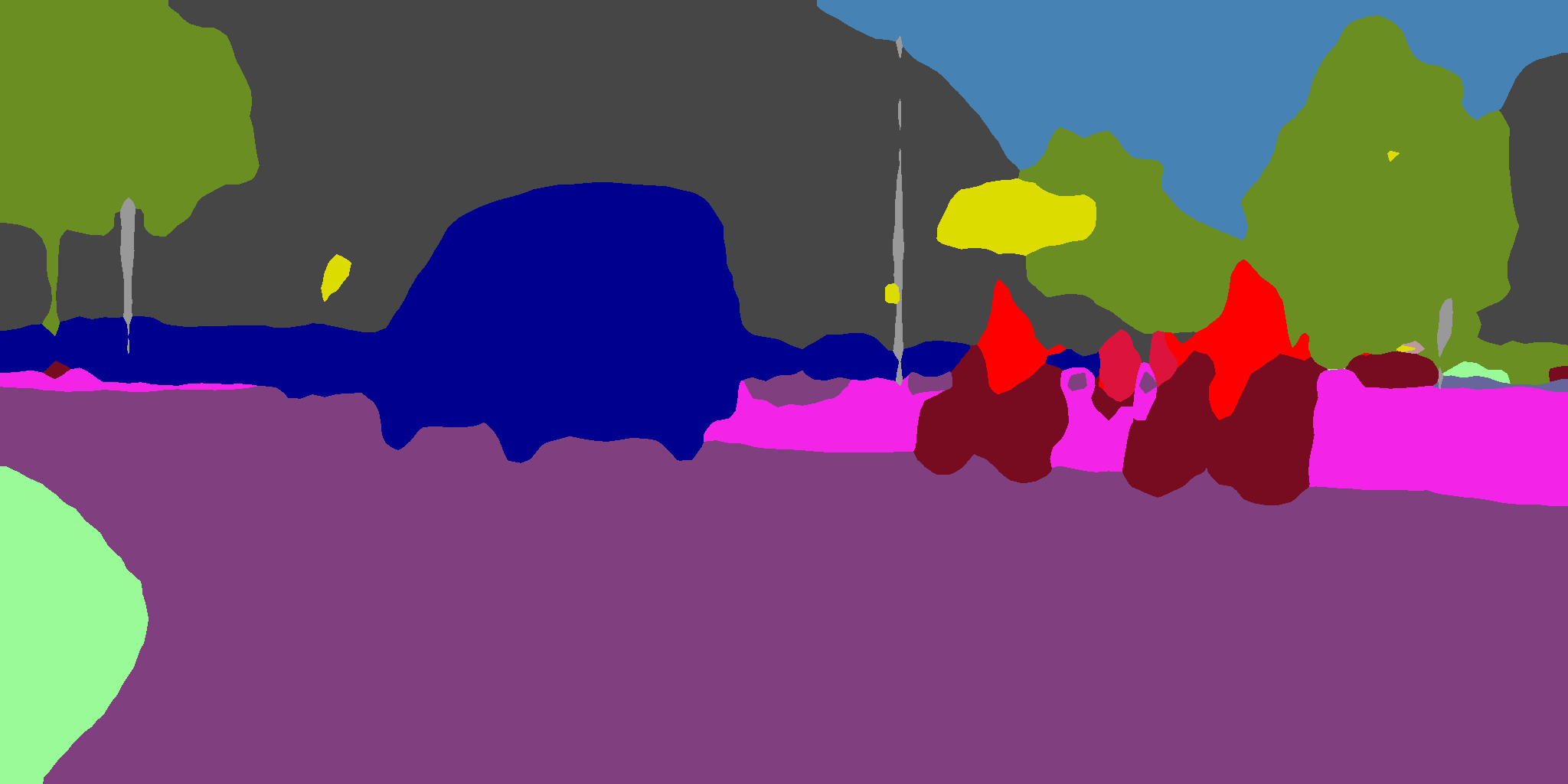}}
	\vspace{+1mm}
	\caption{\textbf{Qualitative outputs}. Two frame sequences at keyframe interval $10$. Column $k + i$ corresponds to the $i$\textsuperscript{th} frame past keyframe $k$. \textbf{First row:} input frames. \textbf{Second row:} Accel $N^R$ branch / DFF \cite{DFF}. \textbf{Third row:} Accel $N^U$ branch / DeepLab-18. \textbf{Fourth row:} Accel-18. Note how Accel both corrects DFF's warping-related distortions in row 2, including the obscured pedestrians (top example) and the distorted vehicles (bottom example), \textit{and} avoids DeepLab's misclassifications in row 3 on the van (top) and vegetation patch (bottom). Column (c) in the bottom example also qualifies as an error case for Accel, as unlike DeepLab, Accel misses the street sign on the right.}%
	\label{fig:qual}%
\end{figure*}

In Figure \ref{fig:qual}, we compare the qualitative performance of DFF (Accel $N^R$), DeepLab (Accel $N^U$), and Accel ($N^R + N^U$) on two sequences of 10 frames (top and bottom).

\section{Conclusion}

Accel is a fast, high-accuracy video segmentation system that utilizes the combined predictive power of two network pathways: (1) a \textit{reference branch} $N^R$ that extracts high-quality features on a reference keyframe, and warps these features forward using incremental optical flow estimates, and (2) an \textit{update branch} $N^U$ that processes the current frame to correct accumulated temporal error in the reference representation. Comprehensive experiments demonstrate a full range of accuracy-inference speed modalities, from a high-throughput version of Accel that is both faster and more accurate than comparable single-frame models to a high-accuracy version that exceeds state-of-the-art. The full ensemble of Accel models consistently outperforms previous work on the problem at all keyframe intervals, while an ablation study demonstrates that Accel makes significant accuracy gains over its individual components. Finally, the Accel architecture is modular and end-to-end trainable, serving as a general example on how to perform dense prediction tasks efficiently on video.

{\small
	\bibliographystyle{ieee}
	\bibliography{egbib}

\begin{thebibliography}{10}\itemsep=-1pt

\bibitem{Pascal}
Y.~Aytar.
\newblock Pascal voc challenge performance evaluation server.
\newblock \url{http://host.robots.ox.ac.uk:8080/leaderboard/}.
\newblock Accessed: 2018-11-07.

\bibitem{SegNet}
V.~Badrinarayanan, A.~Kendall, and R.~Cipolla.
\newblock Segnet: A deep convolutional encoder-decoder architecture for image
  segmentation.
\newblock In {\em PAMI}, 2017.

\bibitem{CamVid}
G.~J. Brostow, J.~Fauqueur, and R.~Cipolla.
\newblock Semantic object classes in video: A high-definition ground truth
  database.
\newblock {\em Pattern Recognition Letters}, 30(2):88–97, 2009.

\bibitem{DeepLabV1}
L.-C. Chen, G.~Papandreou, I.~Kokkinos, K.~Murphy, and A.~L. Yuille.
\newblock Semantic image segmentation with deep convolutional nets and fully
  connected crfs.
\newblock In {\em ICLR}, 2016.

\bibitem{DeepLab}
L.-C. Chen, G.~Papandreou, I.~Kokkinos, K.~Murphy, and A.~L. Yuille.
\newblock Deeplab: Semantic image segmentation with deep convolutional nets,
  atrous convolution, and fully connected crfs.
\newblock In {\em PAMI}, 2017.

\bibitem{MXNet}
T.~Chen, M.~Li, Y.~Li, M.~Lin, N.~Wang, and M.~Wang.
\newblock Mxnet: A flexible and efficient machine learning library for
  heterogeneous distributed systems.
\newblock In {\em NIPS Workshop on Machine Learning Systems}, 2016.

\bibitem{Cityscapes}
M.~Cordts, M.~Omran, S.~Ramos, T.~Rehfeld, M.~Enzweiler, R.~Benenson,
  U.~Franke, S.~Roth, and B.~Schiele.
\newblock The cityscapes dataset for semantic urban scene understanding.
\newblock In {\em CVPR}, 2016.

\bibitem{DCN}
J.~Dai, H.~Qi, Y.~Xiong, Y.~Li, G.~Zhang, H.~Hu, and Y.~Wei.
\newblock Deformable convolutional networks.
\newblock In {\em ICCV}, 2017.

\bibitem{FlowNet}
A.~Dosovitskiy, P.~Fischer, E.~Ilg, P.~H{\"a}usser, C.~Hazrbas, V.~Golkov,
  P.~v.d. Smagt, D.~Cremers, and T.~Brox.
\newblock Flownet: Learning optical flow with convolutional networks.
\newblock In {\em ICCV}, 2015.

\bibitem{PascalVOC}
M.~Everingham, S.~Eslami, L.~V. Gool, C.~Williams, J.~Winn, and A.~Zisserman.
\newblock The pascal visual object classes challenge: A retrospective.
\newblock {\em IJCV}, 111(1):98--136, 2015.

\bibitem{STResNets}
C.~Feichtenhofer, A.~Pinz, and R.~Wildes.
\newblock Spatiotemporal residual networks for video action recognition.
\newblock In {\em NIPS}, 2016.

\bibitem{DeepReps}
C.~Feichtenhofer, A.~Pinz, R.~P. Wildes, and A.~Zisserman.
\newblock What have we learned from deep representations for action
  recognition?
\newblock In {\em CVPR}, 2018.

\bibitem{TwoStream2}
C.~Feichtenhofer, A.~Pinz, and A.~Zisserman.
\newblock Convolutional two-stream network fusion for video action recognition.
\newblock In {\em CVPR}, 2016.

\bibitem{EGB}
P.~Felzenszwalb and D.~Huttenlocher.
\newblock Efficient graph-based image segmentation.
\newblock {\em IJCV}, 59(2):167--181, 2004.

\bibitem{MovingObjects}
K.~Fragkiadaki, P.~Arbelaez, P.~Felsen, and J.~Malik.
\newblock Learning to segment moving objects in videos.
\newblock In {\em CVPR}, 2015.

\bibitem{NetWarp}
R.~Gadde, V.~Jampani, and P.~V. Gehler.
\newblock Semantic video cnns through representation warping.
\newblock In {\em ICCV}, 2017.

\bibitem{KITTI}
A.~Geiger, P.~Lenz, C.~Stiller, and R.~Urtasun.
\newblock Vision meets robotics: The kitti dataset.
\newblock {\em International Journal of Robotics Research (IJRR)}, 32(11),
  2013.

\bibitem{EHG}
M.~Grundmann, V.~Kwatra, M.~Han, and I.~Essa.
\newblock Efficient hierarchical graph-based video segmentation.
\newblock In {\em CVPR}, 2010.

\bibitem{BMV}
S.~Jain and J.~Gonzalez.
\newblock Fast semantic segmentation on video using block motion-based feature
  interpolation.
\newblock In {\em ECCV International Workshop on Video Segmentation}, 2018.

\bibitem{FusionSeg}
S.~D. Jain, B.~Xiong, and K.~Grauman.
\newblock Fusionseg: Learning to combine motion and appearance for fully
  automatic segmentation of generic objects in videos.
\newblock In {\em CVPR}, 2017.

\bibitem{FCDenseNet}
S.~Jégou, M.~Drozdzal, D.~Vazquez, A.~Romero, and Y.~Bengio.
\newblock The one hundred layers tiramisu: Fully convolutional densenets for
  semantic segmentation.
\newblock In {\em CVPR Workshop on Computer Vision in Vehicle Technology},
  2017.

\bibitem{Karpathy}
A.~Karpathy, G.~Toderici, S.~Shetty, T.~Leung, R.~Sukthankar, and L.~Fei-Fei.
\newblock Large-scale video classification with convolutional neural networks.
\newblock In {\em CVPR}, 2014.

\bibitem{GEP}
P.~Kr\"ahenb\"uhl and V.~Koltun.
\newblock Efficient inference in fully connected crfs with gaussian edge
  potentials.
\newblock In {\em NIPS}, 2011.

\bibitem{LowLatency}
Y.~Li, J.~Shi, and D.~Lin.
\newblock Low-latency video semantic segmentation.
\newblock In {\em CVPR}, 2018.

\bibitem{RefineNet}
G.~Lin, A.~Milan, C.~Shen, and I.~Reid.
\newblock Refinenet: Multi-path refinement networks for high-resolution
  semantic segmentation.
\newblock In {\em CVPR}, 2017.

\bibitem{FCN}
J.~Long, E.~Shelhamer, and T.~Darrell.
\newblock Fully convolutional networks for semantic segmentation.
\newblock In {\em CVPR}, 2015.

\bibitem{Budget}
B.~Mahasseni, S.~Todorovic, and A.~Fern.
\newblock Budget-aware deep semantic video segmentation.
\newblock In {\em CVPR}, 2017.

\bibitem{FewStrokes}
N.~S. Nagaraja, F.~R. Schmidt, and T.~Brox.
\newblock Video segmentation with just a few strokes.
\newblock In {\em CVPR}, 2015.

\bibitem{GRFP}
D.~Nilsson and C.~Sminchisescu.
\newblock Semantic video segmentation by gated recurrent flow propagation.
\newblock In {\em CVPR}, 2018.

\bibitem{FastObjSeg}
A.~Papazoglou and V.~Ferrari.
\newblock Fast object segmentation in unconstrained video.
\newblock In {\em ICCV}, 2013.

\bibitem{UNet}
O.~Ronneberger, P.~Fischer, and T.~Brox.
\newblock U-net: Convolutional networks for biomedical image segmentation.
\newblock In {\em MICCAI}, 2015.

\bibitem{CC}
E.~Shelhamer, K.~Rakelly, J.~Hoffman, and T.~Darrell.
\newblock Clockwork convnets for video semantic segmentation.
\newblock In {\em ECCV International Workshop on Video Segmentation}, 2016.

\bibitem{Texton}
J.~Shotton, J.~Winn, C.~Rother, and A.~Criminisi.
\newblock Textonboost for image understanding: Multi-class object recognition
  and segmentation by jointly modeling texture, layout, and context.
\newblock {\em IJCV}, 81(1):2--23, 2009.

\bibitem{TwoStream}
K.~Simonyan and A.~Zisserman.
\newblock Two-stream convolutional networks for action recognition in videos.
\newblock In {\em NIPS}, 2014.

\bibitem{Sturgess}
P.~Sturgess, K.~Alahari, L.~Ladick\'y, and P.~H.~S. Torr.
\newblock Combining appearance and structure from motion features for road
  scene understanding.
\newblock In {\em BMVC}, 2009.

\bibitem{Tokmakov}
P.~Tokmakov, K.~Alahari, and C.~Schmid.
\newblock Learning video object segmentation with visual memory.
\newblock In {\em CVPR}, 2017.

\bibitem{ObjectFlow}
Y.-H. Tsai, M.-H. Yang, and M.~J. Black.
\newblock Video segmentation via object flow.
\newblock In {\em CVPR}, 2016.

\bibitem{Zuxuan}
Z.~Wu, X.~Wang, Y.-G. Jiang, H.~Ye, and X.~Xue.
\newblock Modeling spatial-temporal clues in a hybrid deep learning framework
  for video classification.
\newblock In {\em ACM-MM}, 2015.

\bibitem{DynamicVSN}
Y.-S. Xu, T.-J. Fu, H.-K. Yang, and C.-Y. Lee.
\newblock Dynamic video segmentation network.
\newblock In {\em CVPR}, 2018.

\bibitem{Multiscale}
F.~Yu and V.~Koltun.
\newblock Multi-scale context aggregation by dilated convolutions.
\newblock In {\em ICLR}, 2016.

\bibitem{DRN}
F.~Yu, V.~Koltun, and T.~Funkhouser.
\newblock Dilated residual networks.
\newblock In {\em CVPR}, 2017.

\bibitem{PSPNet}
H.~Zhao, J.~Shi, X.~Qi, X.~Wang, and J.~Jia.
\newblock Pyramid scene parsing network.
\newblock In {\em CVPR}, 2017.

\bibitem{HighPerfVOD}
X.~Zhu, J.~Dai, L.~Yuan, and Y.~Wei.
\newblock Toward high performance video object detection.
\newblock In {\em CVPR}, 2018.

\bibitem{DFF}
X.~Zhu, Y.~Xiong, J.~Dai, L.~Yuan, and Y.~Wei.
\newblock Deep feature flow for video recognition.
\newblock In {\em CVPR}, 2017.

\end{thebibliography}
}

\end{document}